\documentclass[preprint,12pt,authoryear]{elsarticle}

\usepackage[utf8]{inputenc} 
\usepackage[T1]{fontenc}    
\usepackage{hyperref}       
\usepackage{url}            
\usepackage{booktabs}       
\usepackage{amsfonts}       
\usepackage{nicefrac}       
\usepackage{microtype}      
\usepackage[dvipsnames]{xcolor}         

\usepackage{float}
\usepackage{url}
\usepackage{amsmath,amssymb,amsthm}
\usepackage{graphicx}
\usepackage{dsfont}
\usepackage{listings}
\usepackage{colortbl}
\usepackage{natbib}

\usepackage[normalem]{ulem}

\setcitestyle{authoryear,open={(},close={)}}

\newtheorem{definition}{Definition}

\makeatletter
\def\ps@pprintTitle{%
	\let\@oddhead\@empty
	\let\@evenhead\@empty
	\let\@oddfoot\@empty
	\let\@evenfoot\@empty}
\makeatother

\newenvironment{bodyspacing}
{%
	\setlength{\parindent}{0pt}%
	\setlength{\parskip}{0.5em}%
}
{}  

\begin{document}
	
	\begin{frontmatter}
		
		
		
\title{UFO: A Domain-Unification-Free Operator Framework for Generalized Operator Learning}

\author{Hanli Qiao\corref{cor1}\fnref{label1}}
\ead{hanli.qiao@gtk.fi}
\cortext[cor1]{Corresponding author}
\affiliation[label1]{organization={Water and Mining Environment Unit, Geological Survey of Finland},
	            addressline={Vuorimiehentie 5}, 
	            city={Espoo},
	            postcode={02151}, 
	            country={Finland}}	  
          
\author{George~Em~Karniadakis\fnref{label2}}
\affiliation[label2]{organization={Division of Applied Mathematics, Brown University},
	addressline={170 Hope Street}, 
	city={Providence},
	postcode={RI 02912}, 
	country={USA}}

\author{Muhammad~Muniruzzaman\corref{cor1}\fnref{label3}}
\ead{m.muniruzzaman@uni-bonn.de}
\affiliation[label3]{organization={Institute of Geosciences, University of Bonn},
	addressline={Kirschallee 1-3}, 
	city={Bonn},
	postcode={53115}, 
	country={Germany}}

\date{}

\begin{abstract}
Neural operators have become an effective framework for learning mappings between function spaces, yet most existing architectures realize operators within a single representational domain, such as physical, spectral, or latent space. In this work, we introduce UFO (Domain-Unification-Free Operator), a cross-domain neural operator framework that realizes operators through adaptive, jointly conditioned interactions among representations defined on distinct domains. UFO enables discretization decoupling: the input function can be observed at resolutions or locations different from those used during training, while the solution can be queried at arbitrary output resolutions. Across four complementary benchmarks covering discontinuous inputs, irregular sampling with spectral mismatch, nonlinear dynamics, and stochastic high-frequency fields, UFO delivers accurate, robust, and physically coherent predictions under distribution shifts. These results establish cross-domain, phase-modulated realization as a powerful framework for discretization-decoupled neural operator learning.
\end{abstract}

\end{frontmatter}

\begin{bodyspacing}
\section{Introduction}
Learning nonlinear operators that map functions to functions is a central problem in scientific machine learning, with applications spanning fluid mechanics, geophysics, climate modeling, and materials science \citep{WEN2022104180, Pathak2022, Choi2024, Huang2024, Kamyar2024}. Over the past several years, neural operators have emerged as a powerful paradigm for approximating solution operators of parametric partial differential equations (PDEs) and related infinite-dimensional mappings directly from data \citep{Lu2021,Kovachki2023}.

Two families of neural operators have been particularly influential. Deep Operator Networks (DeepONets) represent operators through a decomposition into a branch network, which encodes the input function, and a trunk network, which evaluates the solution at queried spatial or temporal coordinates \citep{Lu2021}. This architecture offers flexibility with respect to irregular geometries, scattered sensors, and physics-informed constraints. On the other hand, Fourier Neural Operators (FNOs) and related spectral architectures embed convolutional kernels in Fourier space, enabling efficient learning of global interactions and improved handling of high-frequency modes \citep{Li2021}.

Although the two families differ in architecture and inductive bias, they are fundamentally realized within a single representational domain. In DeepONet family, operator realization is carried out in the physical domain through branch-trunk representations. In contrast, FNO-type methods realize operators within a spectral domain via global Fourier representations. Despite their success, such single-domain formulations introduce inherent limitations. For instance, DeepONet lacks explicit mechanisms for representing frequency content, often resulting in spectral bias when learning operators associated with highly oscillatory solutions \citep{Rahaman2019,Wang2022}. Meanwhile, FNO-type methods rely on inverse Fourier transforms that typically require regular grids and fixed spectral representations, which may limit their applicability to heterogeneous domains. Another challenge lies in the incorporation of physics-informed constraints, as these often require explicit derivative evaluations in the physical domain \citep{Kovachki2023,Li2024}.

Subsequent works have addressed these difficulties from several complementary directions. Recent studies improve spectral expressivity to better capture high-frequency, oscillatory, or non-smooth solution structures \citep{ZHU2023, Ahmad2026, Khodakarami2026, Cheng2025, Jiang2024, Sojitra2026, Zhao2025}. In parallel, the regular-grid restriction has motivated neural operators for irregular geometries, non-uniform meshes, and flexible input formats. These mechanisms mainly rely on learned geometric deformations, graph- or attention-based representations, and mesh-adaptive constructions \citep{GINO23, GeoFNO24, hao2023, FU2025, Liu_2025_CVPR, Yin2024}. These advances significantly broaden the applicability of neural operators beyond rectangular grids and smooth solution regimes.

Resolution invariance during evaluation is another key motivation of neural operators. However, existing architectures still exhibit practical forms of discretization dependence: DeepONet-type models usually rely on a fixed set of input sensors, while FNO-type models require compatible input-output grid representations. This limits their use in regimes where dense measurements are expensive or unavailable, but high-resolution predictions are required. Several recent works aim to reduce this dependence, including learning operators in latent spaces \citep{LNO2024, Kontolati2024}, resolution-independent neural operators \citep{Bahador2025}, Laplace neural operator \citep{Cao2024}, and super-resolution neural operators \citep{Wei_2023_CVPR}.

Despite these developments, most existing approaches improve expressivity, geometry handling, or resolution transfer through architectural augmentation within a predefined realization mechanism. In particular, representations from different sources or domains are often encoded separately and then combined through fixed fusion rules, such as concatenation, inner products, interpolation, or attention-based aggregation. Such mechanisms enrich the features available to the model, but the operator realization itself remains governed by a fixed composition. 

Motivated by this observation, we introduce UFO (Domain-Unification-Free Operator), a cross-domain operator framework, where the operator is realized through non-separable, jointly conditioned interactions among representations defined on distinct domains. Under this view, operator realization is no longer a fixed mapping, but emerges from an adaptive, phase-modulated coupling between heterogeneous input-space and solution-space representations. 

This work makes three main contributions. First, we introduce a novel operator realization mechanism, UFO. Second, we develop the foundational UFO architecture, consisting of a spectral encoder, a spatial basis network, and an adaptive phase-modulated coupling operator. This architecture enables discretization decoupling, which means that input functions can be observed at resolutions or locations different from those used during training, while solutions can be queried at arbitrary output resolutions. Third, we design a set of complementary benchmarks to evaluate not only pointwise accuracy, but also spectral consistency, structural coherence, generalization, and resolution behavior. 

Specifically, we evaluate UFO on four complementary benchmarks: StepHeat for discontinuous-input spectral bias, Delta-Helmholtz for translation consistency under irregular sampling and spectrum mismatch, Burgers for nonlinear structure preservation under bidirectional extrapolation, and GRF-Helmholtz for stochastic high-frequency random-field operators. Across these settings, UFO delivers accurate, robust, and physically coherent predictions under distribution shifts.

\section{UFO theory and architecture}\label{sec.2}
Let \(\mathcal{A}\) and \(\mathcal{U}\) be Banach spaces of functions defined on domains \(\Omega_{in}\) and \(\Omega_{out}\) , respectively. An operator learning problem consists of approximating a (possibly nonlinear) operator \(\mathcal{G}:\ \mathcal{A}\rightarrow \mathcal{U}\) from a finite set of input-output function pairs \(\{(f_i, \mathcal{G}(f_i))\}\). A neural operator is a parametric family of maps \(\mathcal{G}_\theta\) designed to approximate \(\mathcal{G}\) uniformly on compact subsets of \(\mathcal{A}\).

The solution operator \(\mathcal G\) in UFO is realized through an adaptive, phase-modulated coupling among cross-domain representations, which gives rise to the following definition.

\begin{definition}[Cross-domain Operator Learning]\label{def1}
	A neural operator \(\mathcal{G}_\alpha\) is said to employ cross-domain operator learning if it is realized through a non-separable, jointly-conditioned interaction among representations defined on distinct spaces, i.e.,
	\[
	\mathcal{G}_{\alpha}(f)(x) = \mathcal{C}_{\alpha}(\Phi_{\mathcal{S}}(x),\ \Psi_{\mathcal{H}}(f)),
	\qquad
	x\in \Omega_{out}, 
	\]
where \(\Phi_{\mathcal{S}}(x)\in \mathcal{S},\  \Psi_{\mathcal{H}}(f)\in \mathcal{H}, \  \mathcal{S} \not= \mathcal{H}\), and \(\mathcal{C}_{\alpha}\) is a learnable coupling whose realization depends jointly on these representations and is not reducible to independent mappings acting on each representation separately.\end{definition} 

Clearly, operator realization in Definition \ref{def1} emerges from adaptive interactions among representations defined on distinct spaces. DeepONet-type methods realize operators within a single physical-domain representation, whereas FNO-type methods realize operators within a single spectral-domain representation. Hence, neither family belongs to the cross-domain operator learning (CDOL) framework. 

\paragraph{Remark} The definitions above formalize the operator realization principle introduced by UFO. It naturally admits multi-domain extensions within the same UFO framework.

\subsection{UFO architecture}\label{sec2.1}
As shown in Fig. \ref{Fig1.UFO_archt}, the above principles are instantiated through three tightly coupled modules: the spectral encoder for input-domain representation, the spatial basis network for solution-domain representation, and the adaptive phase-modulated coupling operator for cross-domain realization. Fig. \ref{Fig1.UFO_archt} presents the architecture corresponding to the minimal two-domain form of UFO. 

\begin{figure}[H]
	\centering
	\includegraphics[width=\textwidth]{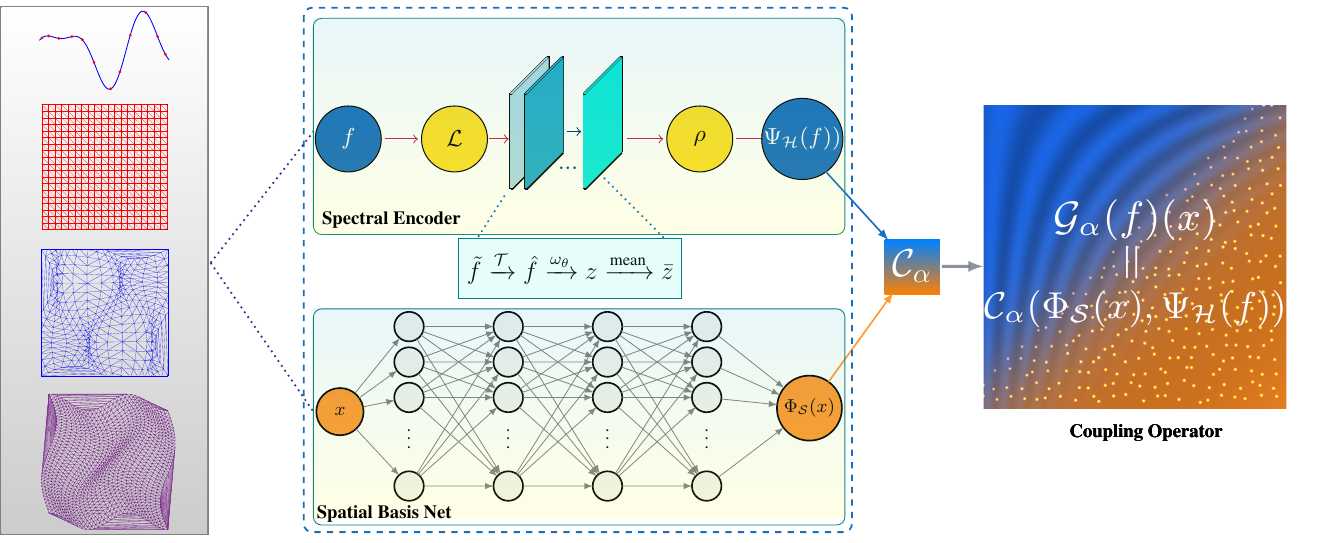}
	\caption{Architecture of the UFO framework.}
	\label{Fig1.UFO_archt}
\end{figure}

\paragraph{Spectral encoder (SE)} The SE is a central component of UFO that constructs a global representation of the input function in a domain distinct from the physical space. To this end, we introduce a learnable, coordinate-conditioned SE that maps an input function to a representation in a spectral domain.

Let \(f:\Omega \rightarrow \mathbb{R}^{d_f}\) be an input function observed at locations \(\{x'_i\}_{i=1}^N \subset \Omega\). The SE constructs the input-domain representation \(\Psi_{\mathcal H}(f) \in \mathbb{C}^{C}\) in UFO. We first lift \(f\) into a higher-dimensional latent space:
\(\tilde f_i = \mathcal L_\theta(f(x'_i)), \  \tilde f_i \in \mathbb{R}^{d_\ell},\)
where \(\mathcal L_\theta\) is a learnable linear lifting map.

A spectral transformation is then applied to the lifted sample sequence, \(\hat f = \mathcal T(\tilde f)\). In our implementation, \(\mathcal{T}\) is instantiated as an FFT-based transform that provides an initial spectral coordinate system for learnable representations, rather than an exact discretization-dependent Fourier expansion. UFO does not rely on any specific choice of transform.

The spectral coefficients are further modulated by a coordinate-conditioned weighting function to enable discretization-agnostic and resolution-invariant representations. Specifically, we introduce \(\omega_\theta : \Omega \rightarrow \mathbb{R}^{d_\ell},\) and for each sampling location \(x'_i\), we perform element-wise modulation:
\[z_i=\omega_\theta(x'_i)\odot\hat f_i.\]
The global representation of \(f\) is then obtained via mean aggregation, \(\displaystyle \bar{z} = \frac{1}{N} \sum_{i=1}^{N} z_i,\) which yields a global spectral summary of the entire input function. In the continuum limit, this operation can be interpreted as a learned spectral integral of the form:
\[\bar z \approx \int_\Omega \omega_\theta(x')\hat f(x')d\mu(x'),\]
capturing the global structure of the function in the spectral domain. Here, \(\mu\) denotes the sampling measure induced by the input discretization.

Finally, we learn a spectral representation through separate nonlinear mappings applied to the real and imaginary components of \(\bar{z}\):
\begin{equation}\label{SE_eq}
	\Psi_{\mathcal H}(f) =
\rho_r(\operatorname{Re}(\bar{z})) + i\,\rho_i(\operatorname{Im}(\bar{z})),
\end{equation}
where \(\rho_r\) and \(\rho_i\) are multilayer perceptrons. This step decouples the learned representation from any specific spectral basis (e.g., Fourier), allowing the model to construct an expressive representation adapted to the operator learning task.

The SE differs fundamentally from existing neural operator constructions. Instead of realizing the operator within a fixed domain (e.g., purely physical or spectral), it constructs a global, coordinate-conditioned representation in a distinct domain. It is subsequently coupled with spatial representations through a non-separable and phase-modulated coupling operator. This design is central to UFO: operator realization emerges from coupling a complementary input-domain representation with spatial representations of the solution space, enabling structured cross-domain interaction.

\paragraph{Spatial basis network} Complementary to the global spectral representation of the input function generated by the SE, UFO constructs a continuous representation of the solution domain through a spatial basis network (SBN). Given query coordinates \(x \in \Omega\), we define \(\Phi_{\mathcal S}(x) \in \mathbb{R}^{C}\) as a learned spatial feature representation. Here, \(x\) is usually different from the observed locations \(x'\) in the SE. Specifically, the SBN is implemented as a multilayer perceptron:
\[\Phi_{\mathcal S}(x) = \phi_\theta(x),\]
where \(\phi_\theta\) maps coordinates directly to a high-dimensional feature space. This formulation is independent of any discretization and naturally supports arbitrary query resolutions and irregular locations. Unlike grid-based representations, the SBN does not construct the solution through convolution or interpolation. Instead, it learns a parameterization of the solution space, providing a continuous spatial basis through which the cross-domain operator realization is carried out.

\paragraph{Adaptive phase-modulated cross-domain coupling operator} With the input function encoded as a global spectral representation \(\Psi_{\mathcal H}(f)\) and the solution domain represented through spatial features \(\Phi_{\mathcal S}(x\)), UFO realizes the operator through a non-separable and phase-modulated cross-domain coupling:
\[\mathcal G_\alpha(f)(x)
=
\mathcal C_\alpha\big(\Phi_{\mathcal S}(x), \Psi_{\mathcal H}(f)\big).\]
The coupling operator \(\mathcal C_\alpha\) is designed to model structured interactions between the two representations. Let \(\Psi_{\mathcal H}(f)  = a + i b, \  a,b \in \mathbb{R}^{C},\) and \(\Phi_{\mathcal S}(x) \in \mathbb{R}^{C}\). We construct a joint feature \(\eta(\Phi_{\mathcal{S}}(x),\ \Psi_{\mathcal{H}}(f)) = \big[\Phi_{\mathcal S}(x),\, a,\, b\big],\) which is used to generate the coupling phase:
\[\alpha = \gamma_\theta\big(\eta(\Phi_{\mathcal{S}}(x),\ \Psi_{\mathcal{H}}(f)) \big),\]
where \(\gamma_\theta\) is a learnable mapping.

The coupling operator is then realized through a phase-modulated interaction, where bounded trigonometric modulation (\(\sin^2\alpha + \cos^2\alpha=1\) ) provides a stable mechanism for structure-aware cross-domain coupling
\begin{align}\label{coupling_eq}
\mathcal G_\alpha(f)(x)
&= \langle \Phi_\mathcal S(x),\, \cos \alpha \odot a + \sin \alpha \odot b\rangle \nonumber \\ 
&= \sum_{c=1}^{C} \big(u_r^{(c)} + u_i^{(c)}\big),
\end{align} 
where \(u_r = \cos(\alpha)\odot a \odot \Phi_{\mathcal S}(x), \  u_i = \sin(\alpha)\odot b \odot \Phi_{\mathcal S}(x).\) Rather than decomposing the operator into independent mappings over input and output domains, UFO realizes it as a jointly conditioned, non-separable, and phase-modulated interaction between heterogeneous representations.

\section{Experiments and results}\label{sec. 3}
We design four benchmarks with complementary roles to evaluate how an operator is realized, along with the predicted performance. \emph{StepHeat} targets spectral bias induced by discontinuous inputs, \emph{\(\delta\)-Helmholtz} tests translation-consistent realization under strong extrapolation, \emph{Burgers} focuses on nonlinear structure preservation, and \emph{GRF-Helmholtz} examines frequency-dominated Gaussian random field (GRF) operators. These problems are chosen to expose distinct regimes in which different operator realizations succeed or degrade.

Accordingly, we report relative \(L^2\) error for pointwise accuracy and Barron norm \citep{Barron2002} relative error, which is quantified between the predicted and reference solutions in a frequency-weighted spectral norm. It is particularly useful for evaluating the model’s ability to capture oscillatory or high-frequency features \citep{Siavash2026}. A larger value indicates that the prediction deviates more from the reference solution in spectral content, suggesting a more inaccurate capture of high-frequency components. Moreover, we analyze qualitative structure through field visualizations and contour alignment. We further distinguish in-distribution (ID) from bidirectional out-of-distribution (OOD) generalization, and examine varying input/output resolutions to test whether the learned operator is tied to a fixed discretization. These evaluations are central to UFO: the goal is not only an accurate approximation, but also a cross-domain operator realization that remains structurally coherent across regimes, resolutions, and distribution shifts.
\subsection{StepHeat: spectral bias under discontinuous inputs}
This benchmark provides a controlled setting to evaluate how operator realization differs between single-domain and cross-domain learning frameworks under spectral bias, particularly in capturing and propagating the high-frequency modes induced by non-smooth inputs.

We consider a one-dimensional heat-type benchmark on \((x,t)\in[0,1]\times[0,1]\) with homogeneous Dirichlet boundary conditions,
\begin{align}
	&u_t=\beta u_{xx}, \nonumber \\
	&u(0,t)=u(1,t)=0,
\end{align}
and a discontinuous step initial condition \(u(x,0)=f_0(x;s)=\mathds{1}_{x>s},\) where \(s\) controls the discontinuity location. 

To control the spectral difficulty, we use the following sine-series realization: \(\displaystyle u(x,t;s)=\sum_{n=1}^N a_n^{(\kappa)}(s)\,
\sin(n\kappa\pi x)\,
e^{-\beta (n\kappa\pi)^2 t},\)
with coefficients
\(a_n^{(\kappa)}(s)
=
\frac{2}{n\kappa\pi}
\Big(\cos(n\kappa\pi s)-\cos(n\kappa\pi)\Big),\)
where \(\kappa\) is a frequency-scaling factor controlling the spectral complexity. When \(\kappa=1\), this reduces to the standard sine-series solution; larger \(\kappa\) yields a more spectrally demanding regime. We set \(\kappa=20\)  and the diffusivity constant \(\beta=6.25\times10^{-4}\). Finally, we  generate 128 training samples by varying the discontinuity location \(s \in [0.3, 0.7]\). 
\begin{table}[H]
	\caption{ID performance on StepHeat with varying discontinuity location \(s\). Relative \(L^2\) and Barron errors are computed over the full spatio-temporal solution field, lower is better.}
	\label{tab1}
	\centering
	\resizebox{\textwidth}{!}
	{
		\begin{tabular}{cccccccc}
			\toprule
			& Metrics & \(s=0.32\) & \(s=0.39\) & \(s=0.41\) & \(s=0.48\) & \(s=0.52\) & \(s=0.66\)\\
			\midrule
			UFO &  & \bf 0.1106 & \bf 0.1166 & \bf 0.1178 & 0.1228 & 0.1267 & \bf 0.0572 \\
			
			DeepONet & Rel. \(L^2\) error & 0.1541 & 0.1888 & 0.2118 & \bf 0.1077 & \bf 0.1078 & 0.0602\\
			
			FNO &  & 0.2356 & 0.3042 & 0.2518 & 0.2407 & 0.2289 & 0.1594\\
			\midrule
			UFO &  & \bf 0.3539 & \bf 0.1934 & \bf 0.1918 & 0.3688 & 0.3731 & 0.3048\\
			
			DeepONet & Barron Norm Rel. error & 0.3459 & 0.2556 & 0.2774 & \bf 0.3046 & \bf 0.3040 & \bf 0.2519\\
			
			FNO &  & 0.6184 & 0.4002 & 0.3420 & 0.6142 & 0.5687 & 0.5723\\
			\bottomrule
	\end{tabular}}
\end{table}
In this benchmark, \(s\) induces distinct spectral patterns, making the problem challenging even in the ID regime as shown in Table \ref{tab1}. This creates a particularly demanding setting for neural operators: DeepONet-type methods are prone to spectral bias under such non-smooth inputs, while FNO-type methods face an additional challenge in handling discontinuities despite their spectral inductive bias. UFO, by contrast, is required to capture and propagate these discontinuity-induced high-frequency modes through cross-domain realization. StepHeat, therefore, serves as a stress test of whether an operator realization can remain accurate and structurally consistent under both non-smooth inputs and strong spectral demands.

Table \ref{tab1} shows clearly that UFO achieves the lowest relative \(L^2\) error on four of the six cases and remains highly competitive on the other two. UFO also attains the best Barron error on the first three cases, while DeepONet is slightly better on the remaining three. This indicates that UFO improves both pointwise reconstruction accuracy and spectral structure preservation in the spatio-temporal solution. 

The competitive performance of DeepONet is consistent with its flexibility in coordinate-based querying. However, the generally higher errors might be an indicator suggesting that physical-domain realization alone is insufficient to fully capture the discontinuity-induced high-frequency modes. FNO performs substantially worse across most cases, with especially large degradation where Barron errors increase sharply. This behavior indicates that, despite its spectral inductive bias, FNO is more sensitive to the discontinuous input family and presents unstable realization quality in the StepHeat benchmark. 

\subsection{\(\delta\)-Helmholtz: translation-consistent realization}
We consider the parametric 2D Helmholtz equation on \(\Omega=[0,1]^2\) with homogeneous Dirichlet boundary conditions,
\begin{equation}\label{helmholtz}
u_{xx}+u_{yy}+k^2u=f,\qquad (x,y)\in\Omega,\quad u|_{\partial\Omega}=0.
\end{equation}
The defined analytical solution is \(u(x,y;\delta;k=10)=(x+y)\sin(10\pi x)\sin(10\pi y)+\delta,\)
where \(\delta\) acts as an additive global shift. The forcing term with mismatch frequency (\(k=1\)) is given by
\(f(x,y;\delta;k=1)
=
\ 2\pi\big(\sin(\pi x)\cos(\pi y)+\cos(\pi x)\sin(\pi y)\big)
+\big(1-2\pi^2\big)(x+y)\sin(\pi x)\sin(\pi y)+\delta.\)

This benchmark is a dual-difficulty task of irregular sampling and cross-frequency, targeting translation-consistent operator realization.  For UFO, each input sample is observed on a randomly and non-uniformly sampled set of locations, and we use regular sampling for DeepONet and FNO due to their architectures. Using 256 samples, we evaluate strong global-shift extrapolation cases to compare the performance of different neural operators under spectral mismatch. 

Fig. \ref{Delta} shows that UFO is the most stable method across both interpolation and extrapolation cases. At \(\delta=4.3\), UFO achieves the lowest relative \(L^2\) and Barron errors, indicating the best performance in both pointwise accuracy and spectral consistency. Under strong extrapolation (\(\delta=\pm 30.8\)), UFO continues to preserve the globally shifted oscillatory structure with mild amplitude deviation, whereas DeepONet exhibits severe structural distortion and FNO collapses under large shifts despite remaining competitive interpolation. Since each UFO input sample is observed on a randomly non-uniform discretization, these results verify that cross-domain realization can maintain translation consistency and stable generalization.
\begin{figure}[H]
	\centering
	\includegraphics[width=0.9\textwidth]{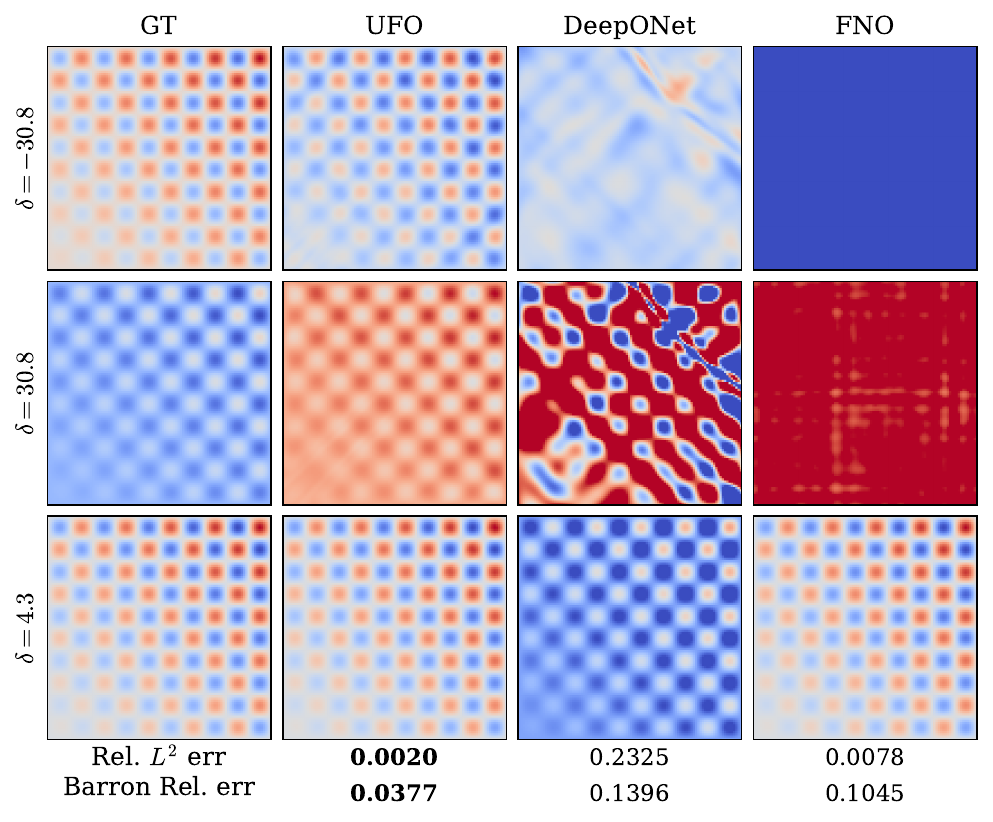}
	\caption{Qualitative comparison on \(\delta\)-Helmholtz for interpolation and extrapolation of global shifts. The model is trained on \(\delta\in[-5,5]\) with 256 samples in total. \(\delta=4.3\) is interpolation, while \(\delta=\pm 30.8\) are strong extrapolation cases. Each UFO sample is observed on a randomly non-uniform input discretization.}
	\label{Delta}
\end{figure}

\subsection{Structure-preserving evaluation on 2D steady Burgers equation}
We further evaluate the models on a 2D steady Burgers equation,
\[
uu_x+uu_y - \nu (u_{xx}+u_{yy}) = f, \quad (x, y) \in [0, 1]^2, \quad u|_{\partial\Omega} = 0
\] 
with a manufactured solution \(u(x,y)=x(1-x)y(1-y)\exp(\lambda(x-y)\)), where \(\nu\) is set to 0.05 and \(\lambda\) controls the deformation of the solution profile, inducing systematic changes in the spatial organization of the field. Models are trained on \(\lambda\in [3,6]\) with 128 samples, and evaluating includes both ID and bidirectional OOD generalization, as shown in Fig. \ref{Burgers} and Figs. \ref{Burgers_UFO} - \ref{Burgers_FNO} (in Appendix). 
\begin{figure}[H]
	\centering
	\includegraphics[width=\textwidth]{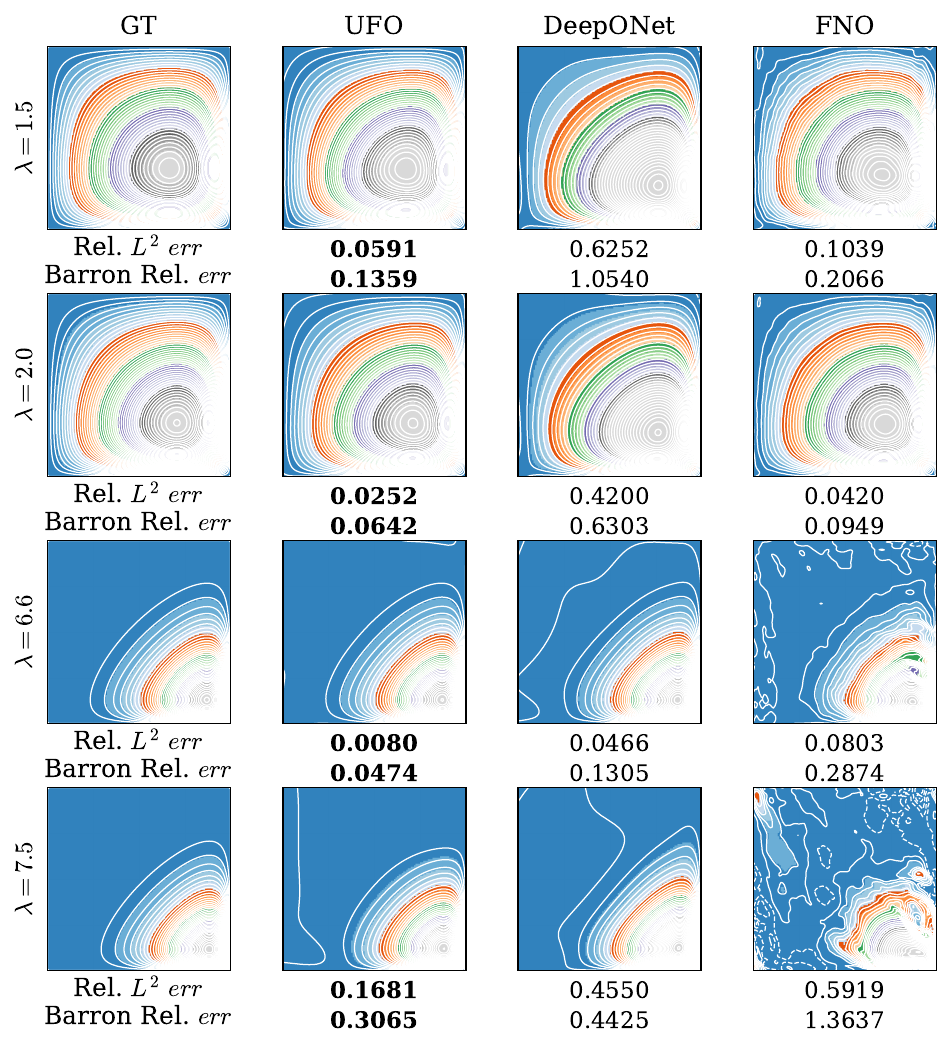}
	\caption{Structural comparison on the parametric 2D Burgers equation under bidirectional OOD generalization. Models are trained on \(\lambda \in [3,6]\). A discretized colormap is used to reveal topological inconsistencies more clearly, where regions belonging to the same value range may become fragmented or disconnected.}
	\label{Burgers}
\end{figure}
In the ID regimes, all methods produce visually accurate solutions (Figs. \ref{Burgers_UFO} - \ref{Burgers_FNO} in Appendix). However, quantitative differences are evident: UFO achieves the lowest errors in relative \(L^2\)  and Barron errors, outperforming both DeepONet and FNO. DeepONet yields stable but consistently higher errors, while FNO exhibits the largest errors despite visually plausible reconstructions.

Under OOD conditions in Fig. \ref{Burgers}, on the left side (\(\lambda=1.5,2.0\)), UFO achieves the lowest relative \(L^2\) and Barron errors, while preserving the contour geometry almost exactly. FNO remains competitive in pointwise accuracy but already exhibits mild topological inconsistency, whereas DeepONet shows clear geometric bias despite producing a smooth field.

The difference becomes more pronounced under right-side extrapolation. At \(\lambda=6.6\) and \(\lambda=7.5\), UFO continues to achieve the lowest relative \(L^2\) and Barron errors while preserving the main nonlinear level-set structure. Unlike the left-side cases, DeepONet becomes more competitive and produces smoother contours. FNO degenerates sharply, with spurious structures and fragmented contours. This indicates that the baselines are sensitive to the extrapolation direction, whereas UFO remains stable across both sides.

\subsection{GRF-Helmholtz: frequency-dominated operator realization}
The last benchmark targets frequency-dominated operator realization on stochastic random field inputs. In contrast to the preceding parameterized benchmarks, which emphasize discontinuities, nonlinear structure, or global shifts, this setting probes whether a neural operator can stably realize solutions when the input is drawn from a spatially correlated GRF family with rich spectral content. 

We consider the 2D Helmholtz equation on \(\Omega=[0,1]^2\) with homogeneous Dirichlet boundary conditions as shown in \eqref{helmholtz} and \(k\) is the wave number. The input forcing \(f(x,y\)) is generated as a spatially correlated GRF using a Mat\'ern kernel with smoothness parameter \((\nu=1.5)\). The full 2D GRF is constructed through a Kronecker-product Cholesky decomposition. Given \(f\), the solution \(u(x,y\)) is computed numerically by finite-difference discretization with sparse matrix formulation.

In the GRF-Helmholtz setting, the model is trained on correlation lengths \(\ell \in \{0.1,0.2,0.3\}\) with 150 training samples each and evaluated at \(\ell=0.05\) and \(\ell=0.35\), corresponding to bidirectional extrapolation toward both shorter and longer correlation scales. 
\begin{figure}[H]
	\centering
	\includegraphics[width=\textwidth]{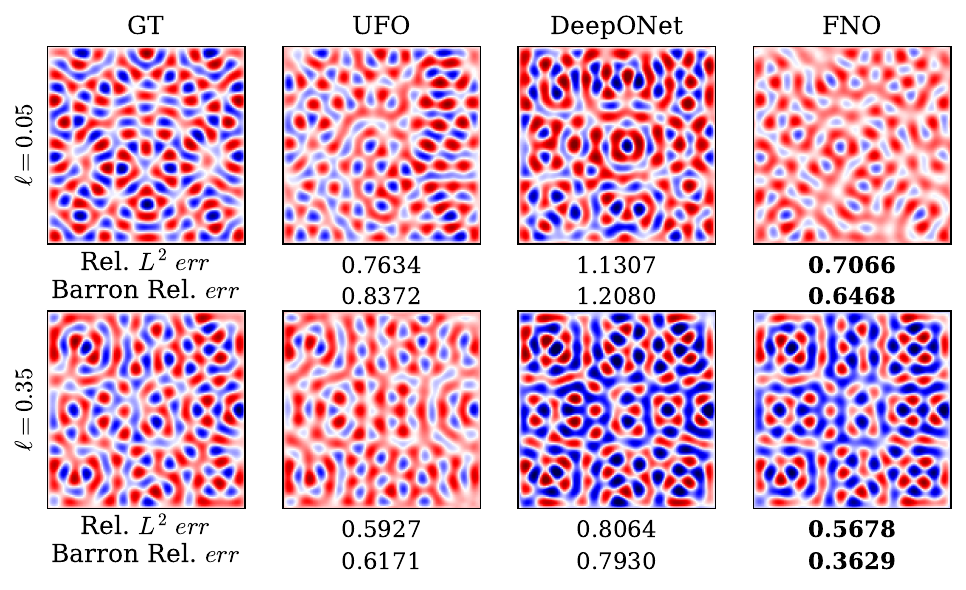}
	\includegraphics[width=\textwidth]{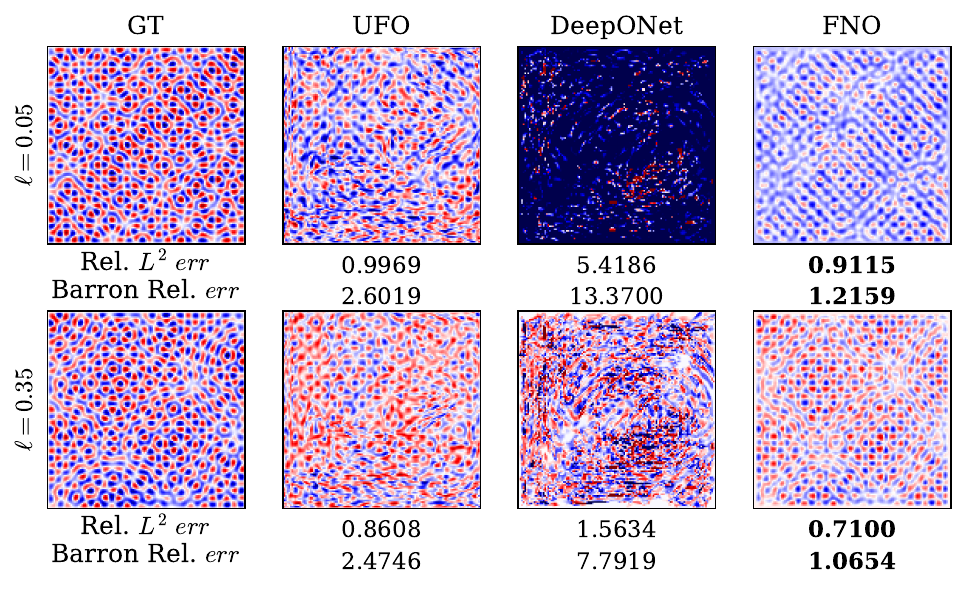}
	\caption{Comparison on GRF-Helmholtz under OOD correlation lengths with moderate wave number \(k=60\) in the top panel and extreme wave number \(k=120\) in the bottom panel.}
	\label{GRFk60k120}
\end{figure}
This testbed is designed to enforce learning from unstructured, non-parametric, and highly oscillatory random inputs. As expected, FNO achieves the best relative \(L^2\) and Barron errors in all cases visualized in Fig. \ref{GRFk60k120}, reflecting its strong inductive bias for frequency-dominated Helmholtz operators. UFO remains consistently competitive and substantially outperforms DeepONet, especially for extreme case \(k=120\), where DeepONet exhibits severe structural distortion and large spectral error.

These results show that GRF-Helmholtz lies close to the natural advantage region of spectral-domain methods. Nevertheless, UFO preserves the dominant random-field structures across the OOD correlation lengths under both wave-number regimes. Together with the stronger results on StepHeat, \(\delta\)-Helmholtz, and Burgers, this suggests that UFO provides a more balanced operator realization mechanism across heterogeneous regimes, rather than specializing only to spectral-domain problems.

\subsection{Discretization-decoupled operator realization}
A key theoretical advantage of UFO is discretization decoupling: the input function can be observed at resolutions or locations different from those used during training, while the solution can be queried at arbitrary output resolutions. 

We evaluate this property from two complementary perspectives that align with the practical concerns. First, we decrease the resolution at which the input function is observed while keeping the output evaluation grid fixed, testing whether the learned operator remains stable when the input discretization differs from training. Second, we fix the input observation and increase the output query resolution, testing whether dense solution fields can be produced without requiring equally dense input observations. 

The top row in Fig. \ref{Resol} varies the input resolution while keeping the output evaluation fixed. Across all tested \(\lambda\) values, UFO exhibits only mild changes in relative \(L^2\) and Barron errors as the input resolution decreases from \(100\times100\) to \(55\times55\). This indicates that the learned input-domain representation is not tightly coupled to a fixed input discretization. The degradation is most visible when the input becomes sparse and the solution regime is more difficult, but the overall error curves remain stable, showing that UFO preserves its operator realization under input resolution changes.

The second row further tests output query resolution by increasing the evaluation grid from \(100\times100\) to \(550\times550\) at \(\lambda=5.8\). UFO remains nearly flat in both relative \(L^2\) and Barron errors, demonstrating that the predicted solution can be queried at much denser resolutions without loss of accuracy. DeepONet also remains relatively stable, but its error is consistently higher than UFO. In contrast, FNO deteriorates rapidly as the output resolution increases, especially in Barron error, reflecting its dependence on grid-tied spectral realization. These results support the key advantage of UFO: the input representation and solution query space are decoupled, enabling stable operator realization across varying input and output resolutions.
\begin{figure}[H]
	\centering
	\includegraphics[width=\textwidth]{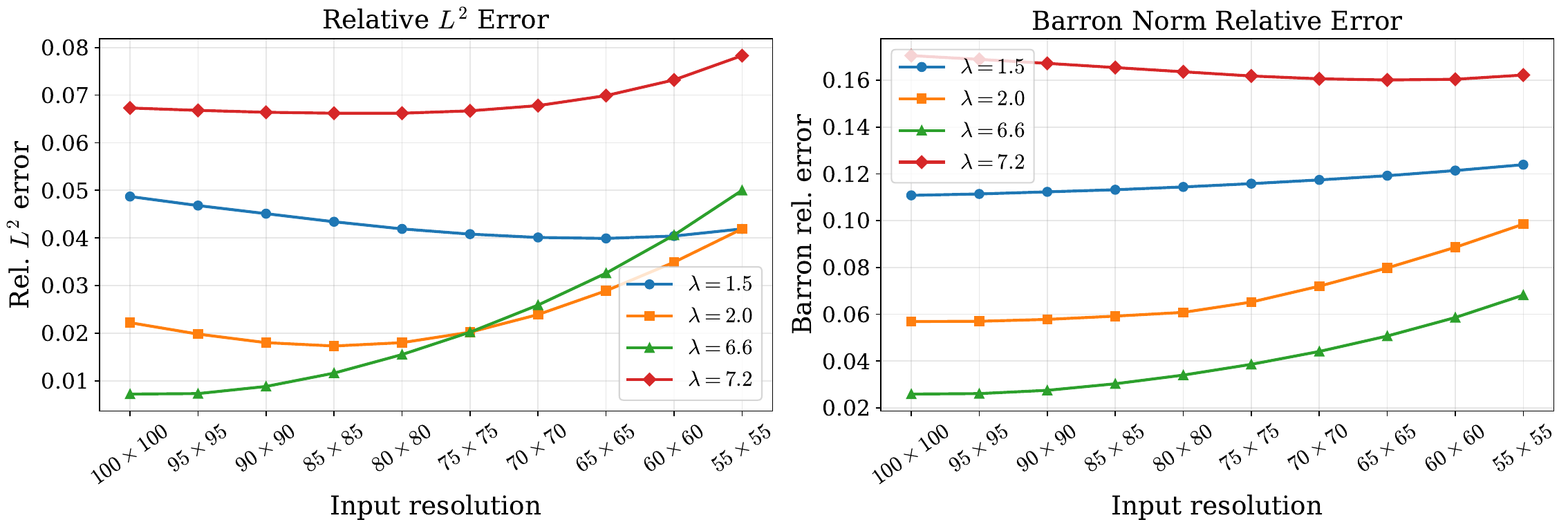}
	\includegraphics[width=\textwidth]{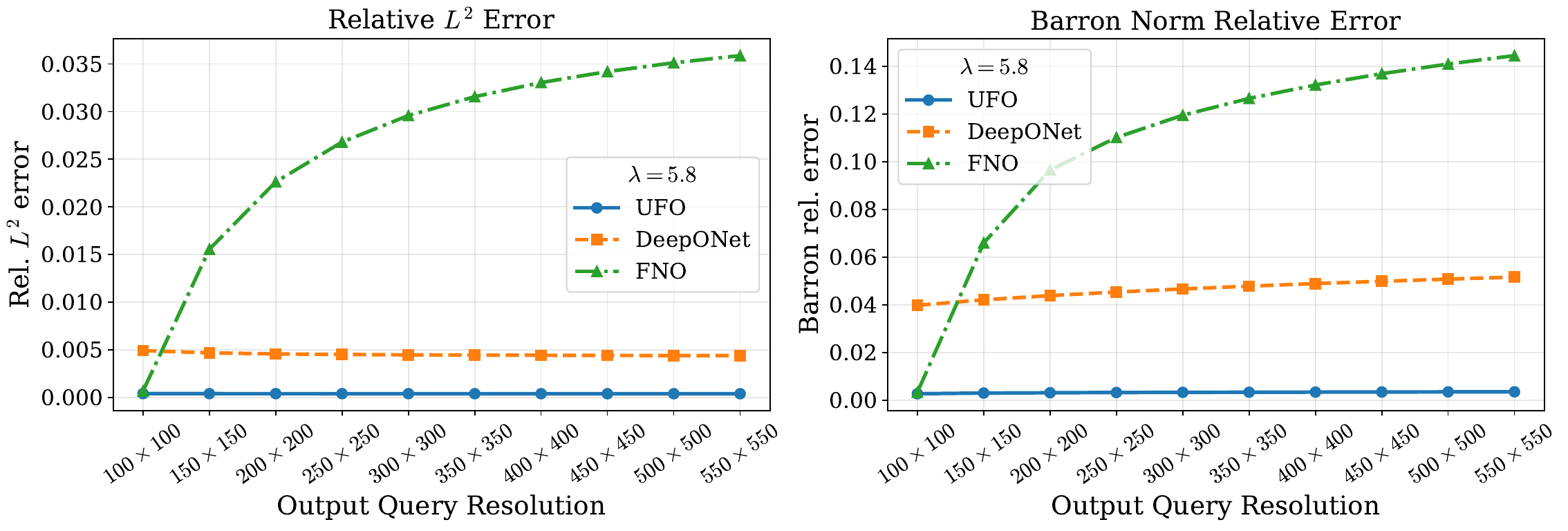}
	\caption{Resolution behavior of UFO on Burgers. Top row: relative \(L^2\) and Barron errors under decreasing input resolutions for different \(\lambda\) values. Second row: output query resolution study at \(\lambda=5.8\).}
	\label{Resol}
\end{figure}

\subsection{Ablation study in StepHeat}
To isolate the role of adaptive phase modulation coupling \(\mathcal{C}_\alpha\), we consider a separable variant of UFO that keeps the SE and SBN unchanged, but replaces the phase-modulated coupling with a fixed separable readout,
\[\mathcal G(f)(x)
=
\langle \Phi_{\mathcal S}(x), a + b\rangle,
\qquad
\Psi_{\mathcal H}(f)=a+ib.\]
This variant preserves the two-domain representation structure of UFO, but removes the jointly conditioned interaction, \(\alpha\), between domains. Therefore, the comparison tests whether UFO’s performance comes merely from having input-domain and solution-domain representations, or from the adaptive, jointly dependent interaction between them.
\begin{table}[H]
	\centering
	\caption{Ablation of the adaptive phase \(\alpha\) on StepHeat. The ablated variant keeps the spectral encoder and spatial basis network unchanged, but removes the jointly conditioned phase modulation and replaces Eq. \eqref{coupling_eq} with a fixed separable readout. Across all discontinuity locations, removing \(\alpha\) substantially degrades both relative \(L^2\) and Barron relative errors, showing that the adaptive phase is essential for coupling the input-domain spectral representation with the solution-domain basis in this non-smooth, high-frequency regime.}
	\label{tab:ablation}
	\resizebox{\textwidth}{!}{
		\begin{tabular}{ccccccccccccc}
			\toprule
			
			& \multicolumn{2}{c}{\(s=0.32\)} 
			& \multicolumn{2}{c}{\(s=0.39\)}  
			& \multicolumn{2}{c}{\(s=0.41\)}
			& \multicolumn{2}{c}{\(s=0.48\)} 
			& \multicolumn{2}{c}{\(s=0.52\)}  
			& \multicolumn{2}{c}{\(s=0.66\)}\\
			\cmidrule(r){2-13}
			& Rel. \(L^2\)  & Barron Rel. & Rel. \(L^2\)  & Barron Rel. & Rel. \(L^2\)  & Barron Rel. 
			& Rel. \(L^2\)  & Barron Rel. & Rel. \(L^2\)  & Barron Rel. & Rel. \(L^2\)  & Barron Rel. \\
			\midrule			
			UFO & 0.1106 & 0.3539 & 0.1166 & 0.1934 & 0.1178 & 0.1918 & 0.1228 & 0.3688 & 0.1267 & 0.3731 & 0.0572 & 0.3048\\
			UFO w/o \(\alpha\)& 0.3365 & 0.4618 & 2.3009 & 0.7179 & 2.3774 & 0.7303 & 0.4098 & 0.4865 & 0.3446 & 0.4659 & 0.4950 & 0.4951\\
			\bottomrule
		\end{tabular}
	}
\end{table}
Table \ref{tab:ablation} isolates the role of the adaptive phase \(\alpha\). Although the model still uses two-domain representations, its errors increase substantially across all StepHeat cases, especially at \(s=0.39\) and \(s=0.41\). The consistent degradation in both relative \(L^2\) and Barron errors shows that \(\alpha\) is essential for adaptive cross-domain realization in the discontinuous, high-frequency regime. We also observe that removing \(\alpha\) makes optimization substantially harder, with training losses decreasing slowly and often failing to converge to the same level as the full UFO.

This confirms that \(\mathcal{C}_\alpha\) is not a cosmetic module. It is the mechanism that turns separate domain representations into an adaptive cross-domain operator realization. Without it, UFO degenerates to a separable multi-domain representation, which lacks the jointly conditioned interaction needed to preserve solution structure under distribution shift.

\section{Discussion}
By separating input-domain encoding, solution-domain representation, and adaptive phase-modulated coupling, UFO reframes neural operator learning from single-domain approximation to cross-domain realization. UFO provides an effective way to realize operators across heterogeneous regimes, including discontinuous inputs, irregular observations, nonlinear dynamics, and frequency-dominated random fields. 

A key implication of UFO is discretization decoupling, which means that the input function can be observed at resolutions or locations different from those used during training, and the output solution queried at arbitrary output resolutions. This is valuable in practical settings where dense measurements are expensive, but high-resolution predictions are required. However, this flexibility is not a free lunch. Specifically, when the target operator relies strongly on localized input features, overly sparse observations will degrade the prediction. Thus, UFO is discretization-flexible, but the input observations must still resolve the structures that determine the operator response.

UFO is not merely a neural operator architecture, but a way to think about operator realization beyond domain constraints. This view opens the door to multi-domain, time-dependent, physics-informed, and multimodal extensions for complex scientific systems. Specifically, it may extend the instantiated minimal two-domain form of UFO via additional domain incorporations, including image-domain representations, multiple spectral transforms, and multiscale physical descriptors. Such extensions are particularly relevant for complex systems, including CFD, reactive transport modeling, inverse problem, and coupled multiphysics dynamics. 

\section*{Acknowledgments}
This study is funded by Jane and Aatos Erkko Foundation via the project titled ML‐Mining: Machine Learning surrogate modeling for risk assessment and water quality prediction at Mining sites (Grant 220021). This work has also been supported by the Research Council of Finland project 372518. The authors also acknowledge the support from the IT Center for Science, Finland (CSC), for generously sharing their computational resources.
\end{bodyspacing}

{\small  \bibliography{UFO_Ref}}


\begin{bodyspacing}
\appendix
\section{Appendix}
\subsection{Table of parameters}
Table \ref{tab2} reports parameter counts and per-epoch runtime. UFO remains compact, with fewer than \(8\times10^4\) trainable parameters in all settings, and achieves competitive runtime compared with the baselines. It is consistently faster than FNO except on Burgers, while the additional cost relative to DeepONet reflects the spectral encoder and adaptive phase-modulated coupling. Thus, UFO’s performance is not driven by model size, but by its cross-domain realization mechanism.
\begin{table}[H]
	\caption{Trainable parameters and per-epoch runtime across benchmarks. UFO uses fewer than \(8\times10^4\) parameters in all settings and maintains competitive runtime, while DeepONet and FNO often require larger parameter counts or higher per-epoch cost.}
	\label{tab2}
	\centering
	\resizebox{\textwidth}{!}
	{
		\begin{tabular}{cccccc}
			\toprule
			 & StepHeat & \(\delta\)-Helmholtz & Burgers & GRF-Helmholtz (\(k=60\)) & GRF-Helmholtz (\(k=120\)) \\
			\midrule
			UFO &  48,442 & 44,332 & 73,452 & 56,762 & 51,734 \\
			DeepONet & 27,456 & 181,056 & 661,056 & 181,056 & 181,056\\
			FNO &  8,422,209 & 8,422,209 & 2,118,273 & 18,907,905 & 8,422,209\\
			\bottomrule
			\multicolumn{6}{c}{Runtime per epoch}\\
			\midrule
			UFO &  0.087\(s\)  &  0.143\(s\) & 0.145\(s\)  & 0.257\(s\)  &  0.228\(s\) \\
			DeepONet & 0.048\(s\)  & 0.050\(s\)  &  0.028\(s\) & 0.065\(s\)  & 0.065\(s\) \\
			FNO &  0.126\(s\)  &  0.449\(s\) &  0.090\(s\) &  0.334\(s\) & 0.265\(s\) \\
			\bottomrule
	\end{tabular}}
\end{table}

\subsection{Benchmarks summary}
\begin{table}[H]\footnotesize
	\centering
	\caption{Summary of benchmarks used in Section \ref{sec. 3}. Each benchmark is designed to probe the operator realization from distinct views, including spectral bias, irregular input sampling, nonlinear structure preservation, stochastic high-frequency fields, and discretization behavior.}
	\label{tab:benchmark_summary}
	\resizebox{\textwidth}{!}{
		\begin{tabular}{p{2.0cm} p{3.8cm} p{3.0cm} p{2.2cm} p{2.2cm} p{4.8cm}}
			\toprule
			\textbf{Benchmark} 
			& \textbf{Governing problem / construction} 
			& \textbf{Purpose} 
			& \textbf{Training distribution} 
			& \textbf{Test setting} 
			& \textbf{Training data generation} \\
			\midrule
			
			StepHeat 
			& 1D heat-type equation with discontinuous step initial condition and high frequency-scaled sine-series solution. 
			& Tests discontinuous-input spectral bias and propagation of high-frequency modes induced by varying jump locations. 
			& $s \in [0.3,0.7]$; $128$ training samples; $\kappa=20$ and diffusivity $\beta=6.25\times10^{-4}$. 
			& ID: $s=$ 0.32, 0.39, 0.41, 0.48, 0.52, 0.66. 
			& Training samples are generated by varying the discontinuity location \(s\). Input step function \(f\) is uniformly observed at 100 locations in \([0,1]\). The query coordinates for the solution field are sampled on a uniform \(150\times50\) space-time grid, with target values generated analytically from the sine-series realization.\\
			
			\midrule
			
			$\delta$-Helmholtz 
			& 2D Helmholtz equation with manufactured solution 
			$u(x,y;\delta)=(x+y)\sin(10\pi x)\sin(10\pi y)+\delta$. 
			& Tests translation-consistent realization under global shifts, irregular input sampling, and spectrum mismatch between forcing (\(k=1\)) and solution (\(k=10\)). 
			& $\delta \in [-5,5]$; $256$ training samples.  
			& ID: $\delta=4.3$; strong bidirectional extrapolation: $\delta=\pm30.8$. 
			& Data are generated from a manufactured analytical solution, with forcing obtained from the Helmholtz operator under frequency mismatch. Training samples are generated by varying \(\delta\). For UFO, each input \(f\) is observed at \(50\times50\) randomly sampled non-uniform spatial locations \((x,y)\), while the solution field is sampled on a uniform \(60\times60\) grid.\\
			
			\midrule
			
			2D Burgers 
			& Steady 2D Burgers equation with manufactured solution 
			$u(x,y)=x(1-x)y(1-y)\exp(\lambda(x-y))$. 
			& Tests nonlinear structure preservation, contour connectivity, and bidirectional OOD extrapolation. 
			& $\lambda \in [3,6]$; $128$ training samples; viscosity $\nu=0.05$. 
			& ID: $\lambda=3.2,3.8,4.2,5.8$; OOD: $\lambda=1.5,2.0,6.6,7.2$. 
			& Training samples are generated from a manufactured solution, with the forcing computed from the Burgers equation. Both the input forcing \(f\) and the output solution field are sampled on a uniform \(100\times100\) spatial grid over \((x,y)\). \\
			
			\midrule
			
			GRF-Helmholtz 
			& 2D Helmholtz equation with stochastic GRF forcing generated from a Matérn covariance kernel. 
			& Tests stochastic high-frequency operator realization and performance in a frequency-dominated regime favorable to spectral methods. 
			& Correlation lengths $\ell\in\{0.1,0.2,0.3\}$; $150$ GRF samples per $\ell$; wave numbers $k=60,120$. 
			& OOD correlation lengths $\ell=0.05$ and $\ell=0.35$. Solutions are evaluated on uniform \(128\times128\) spatial grids over \((x,y)\).
			& GRF forcings are generated by Kronecker-product Cholesky decomposition, with solutions computed by finite-difference discretization and sparse linear solves. Both training \(f\) and solution fields are sampled on a uniform \(50\times50\) spatial grid.\\
			
			\midrule
			
			Resolution study 
			& Burgers-based evaluation with varied input observation resolution and output query resolution. 
			& Tests discretization decoupling from both input-observation and output-query sides. 
			& Models trained with $100\times100$ input observations. 
			& Input resolution varied from $100\times100$ to $55\times55$; output query resolution varied from $100\times100$ to $550\times550$. 
			& Uses trained Burgers models; evaluation only, no additional training. \\
			
			\bottomrule
		\end{tabular}
	}
\end{table}

\subsection{Multi-seed quantitative results}
\begin{table}[H]
	\centering
	\caption{Multi-seed quantitative results across benchmarks. We repeat each experiment with five independent training seeds \(\{42,200,500,2010,2026\}\). The generated datasets and evaluation samples are fixed across seeds, while model initialization and training stochasticity are varied. We report mean \(\pm\) standard deviation for relative \(L^2\) error and Barron norm relative error. Bold values indicate the best mean performance for each metric.}
	\label{tab:seed}
	\resizebox{\textwidth}{!}{
		\begin{tabular}{ccccccc}
			\toprule
			Benchmark
			& \multicolumn{2}{c}{UFO} 
			& \multicolumn{2}{c}{DeepONet}  
			& \multicolumn{2}{c}{FNO}\\
			\cmidrule(r){2-7}
			& Rel. \(L^2\)  & Barron Rel. & Rel. \(L^2\)  & Barron Rel. & Rel. \(L^2\)  & Barron Rel. \\
			\midrule
			
			StepHeat &&&&&& \vspace{0.2cm}\\
			\(s=0.32\) & {\bf 0.1070} \(\pm\) 0.0068 & 0.3413 \(\pm\) 0.0213 & 0.1256 \(\pm\) 0.0180 &
			{\bf 0.3186} \(\pm\) 0.0226 & 0.2360 \(\pm\) 0.0030 &
			0.6248 \(\pm\) 0.0102\\
			\(s=0.39\) & {\bf 0.1090} \(\pm\) 0.0072 & {\bf 0.1734} \(\pm\) 0.0137 & 0.2072 \(\pm\) 0.0176 &
			0.2491 \(\pm\) 0.0070 & 0.3210 \(\pm\) 0.0380 &
			0.4381 \(\pm\) 0.0712\\
			\(s=0.41\) & {\bf 0.1117} \(\pm\) 0.0098 & {\bf 0.1835} \(\pm\) 0.0158 & 0.2434 \(\pm\) 0.0222 &
			0.2801 \(\pm\) 0.0050 &0.3224 \(\pm\) 0.0760 &
			0.4613 \(\pm\) 0.1323\\
			\(s=0.48\) & 0.1078 \(\pm\) 0.0103 & 0.3431 \(\pm\) 0.0267 & {\bf 0.1026} \(\pm\) 0.0047 &
			{\bf 0.2837} \(\pm\) 0.0218 & 0.2276 \(\pm\) 0.0092 &
			0.5682 \(\pm\) 0.0352\\
			\(s=0.52\) & 0.1083 \(\pm\) 0.0110 & 0.3436 \(\pm\) 0.0251 & {\bf 0.1028} \(\pm\) 0.0048 &
			{\bf 0.2834} \(\pm\) 0.0219 & 0.2231 \(\pm\) 0.0068 &
			0.5549 \(\pm\) 0.0378\\
			\(s=0.66\) & 0.0534 \(\pm\) 0.0043 & 0.2875 \(\pm\) 0.0180 & {\bf 0.0533} \(\pm\) 0.0049 &
			{\bf 0.2384} \(\pm\) 0.0123 & 0.1605 \(\pm\) 0.0041 &	0.5836 \(\pm\) 0.0250\\
			\midrule
			
			$\delta$-Helmholtz &&&&&& \vspace{0.2cm}\\
			\(\delta=-30.8\) & {\bf 0.0569} \(\pm\) 0.0501 &
			1.1394 \(\pm\) 1.3670 & 0.0641 \(\pm\) 0.0498 &
			{\bf 0.5667} \(\pm\) 0.2920 & 1.1290 \(\pm\) 1.6856 &
			1.6335 \(\pm\) 1.8539\\
			\(\delta=30.8\) & {\bf 0.0579} \(\pm\) 0.0432 & 
			{\bf 0.4856} \(\pm\) 0.5358 & 0.2222 \(\pm\) 0.0615 & 
			3.2333 \(\pm\) 2.6625 & 0.3899 \(\pm\) 0.3885 &
			0.7637 \(\pm\) 0.6656\\
			\(\delta=4.3\) & {\bf 0.0013} \(\pm\) 0.0006 &
			{\bf 0.0343} \(\pm\) 0.0261 & 0.2279 \(\pm\) 0.0044 &
			0.1331 \(\pm\) 0.0046 & 0.0078 \(\pm\) 0.0000 &	0.1042 \(\pm\) 0.0003\\			
			\midrule
			
			2D Burgers &&&&&& \vspace{0.2cm}\\
			\(\lambda=1.5\) &{\bf 0.0676} \(\pm\) 0.0134 &{\bf0.1612} \(\pm\) 0.0477 &0.5387 \(\pm\) 0.1388 & 0.9144 \(\pm\) 0.2288&0.0782 \(\pm\) 0.0236 & 0.1886 \(\pm\) 0.0240\\
			\(\lambda=2.0\) &{\bf 0.0292} \(\pm\) 0.0077 & {\bf 0.0842} \(\pm\) 0.0305&0.4314 \(\pm\) 0.0255 & 0.6545 \(\pm\) 0.0489&0.0321 \(\pm\) 0.0101 & 0.0923 \(\pm\) 0.0100\\
			\(\lambda=4.5\) &{\bf 0.0024} \(\pm\) 0.0020 & {\bf 0.0083} \(\pm\) 0.0055&0.0074 \(\pm\) 0.0036 & 0.0524 \(\pm\) 0.0120&0.0073 \(\pm\) 0.0005 & 0.0238 \(\pm\) 0.0006\\
			\(\lambda=6.6\) &0.0420 \(\pm\) 0.0224 & {\bf 0.0648} \(\pm\) 0.0182&{\bf 0.0402} \(\pm\) 0.0239 & 0.1282 \(\pm\) 0.0337&0.0840 \(\pm\) 0.0118 & 0.2643 \(\pm\) 0.0343\\
			\(\lambda=7.5\) &{\bf 0.1404} \(\pm\) 0.0681 & 0.4023 \(\pm\) 0.1923&0.3052 \(\pm\) 0.1552 & {\bf 0.4008} \(\pm\) 0.2272&0.6383 \(\pm\) 0.1001 & 1.3082 \(\pm\) 0.1478\\
			\midrule
			
			GRF-Helmholtz &&&&&& \vspace{0.2cm}\\
			\(k=60,\ \ell=0.05\) & 1.0486 \(\pm\) 0.2651 &
			1.0190 \(\pm\) 0.2181 & 1.2074 \(\pm\) 0.1224 &
			1.2996 \(\pm\) 0.1347 & {\bf 0.8428} \(\pm\) 0.0874 &
			{\bf 0.8700} \(\pm\) 0.1289\\
			\(k=60,\ \ell=0.35\) & 1.0027 \(\pm\) 0.3861 &
			1.0747 \(\pm\) 0.4027 & 0.7120 \(\pm\) 0.1222 &
			0.6785 \(\pm\) 0.1004 & {\bf 0.6176} \(\pm\) 0.0283 &
			{\bf 0.5706} \(\pm\) 0.1164\\
			\(k=120,\ \ell=0.05\) &  1.2138 \(\pm\) 0.1931 &
			2.6334 \(\pm\) 0.3025 & 2.9746 \(\pm\) 1.6753 &
			11.3943 \(\pm\) 5.0178 & {\bf 0.9096} \(\pm\) 0.0238 &
			{\bf 1.2323 }\(\pm\) 0.0377\\
			\(k=120,\ \ell=0.35\) & 0.9989 \(\pm\) 0.2265 &
			2.7161 \(\pm\) 0.3049 & 1.5353 \(\pm\) 0.3990 &
			6.8914 \(\pm\) 3.3888 & {\bf 0.7084} \(\pm\) 0.0034 & {\bf 1.0692} \(\pm\) 0.0070\\
			\bottomrule
		\end{tabular}
	}
\end{table}

\subsection{Additional analysis of input observation resolution}
\begin{figure}[H]
	\centering
	\includegraphics[width=0.6\textwidth]{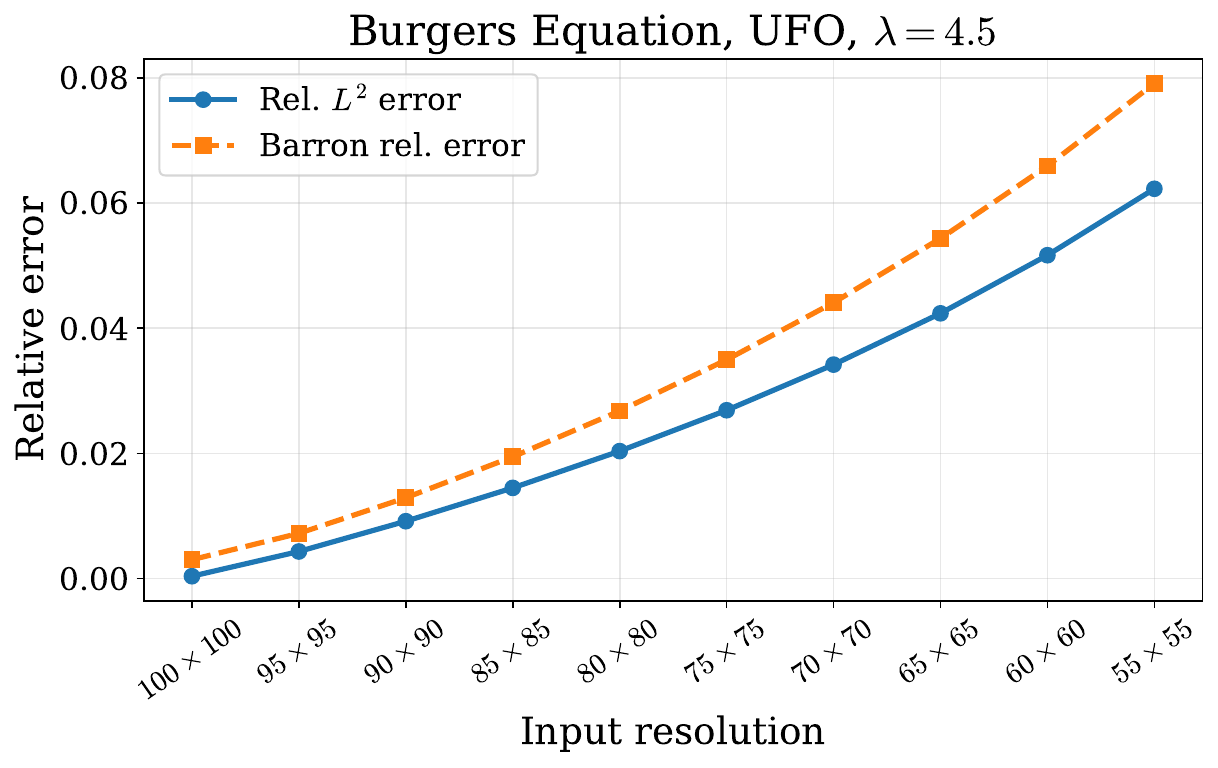}
	\caption{In-distribution input-resolution analysis of UFO on the 2D Burgers equation. UFO is trained with \(100\times100\) input observations and evaluated at \(\lambda=4.5\) using progressively coarser input resolutions from \(100\times100\) to \(55\times55\). The output query resolution is fixed. Unlike the OOD setting in Fig. \ref{Resol}, where UFO remains largely stable under input sparsification, the in-distribution error increases smoothly as input observations become coarser, reflecting the benefit of dense input information for fine-grained ID reconstruction.}
	\label{UFO_ID_inres}
\end{figure}
Fig. \ref{UFO_ID_inres} complements the main resolution-decoupling study. While Fig. \ref{Resol} shows that UFO remains largely stable under input sparsification in the OOD regime, this ID test at \(\lambda=4.5\) reveals a smooth increase in both relative \(L^2\) and Barron errors as the input observation resolution decreases. This does not contradict discretization decoupling: UFO can process input observations at resolutions different from training without requiring a fixed grid, but denser observations still provide more information for fine-grained ID reconstruction. Thus, UFO is discretization-decoupled rather than information-free, and its accuracy degrades gradually when input information is reduced.

\subsection{In distribution predictions on 2D steady Burgers equation}
Figs \ref{Burgers_UFO}-\ref{Burgers_FNO} provide additional in-distribution results on the parametric 2D Burgers equation. Models are trained on \(\lambda\in[3,6]\) and evaluated at \(\lambda=3.2,3.8,4.2,5.8\). The first row shows the ground truth, the second row shows the prediction, and the third row visualizes the absolute error. The two numbers below each column report relative errors of \(L^2\) and Barron norm, respectively. 

All three models recover the main solution structure within the training range, but their error patterns differ substantially. UFO achieves the lowest relative errors of \(L^2\) and Barron norm across all tested \(\lambda\) values, with relative \(L^2\) errors on the order of \(10^{-4}\) and Barron errors below \(8\times10^{-3}\). Its absolute error remains localized and low-magnitude, indicating that the learned operator preserves both pointwise accuracy and the spectral structure of the solution.

DeepONet produces visually plausible fields, but its errors are consistently larger than those of UFO, especially in Barron error. This suggests that although coordinate-based querying captures the global solution geometry, the learned representation is less accurate in preserving spectral content. FNO also reconstructs the coarse solution structure, but exhibits larger error near boundary and high-gradient regions, and its Barron error is substantially higher than UFO. Overall, the in-distribution results confirm that UFO’s advantage is not limited to OOD extrapolation: even within the training regime, it provides more accurate and spectrally consistent operator realization. 
\begin{figure}[H]
	\centering
	\includegraphics[width=0.8\textwidth]{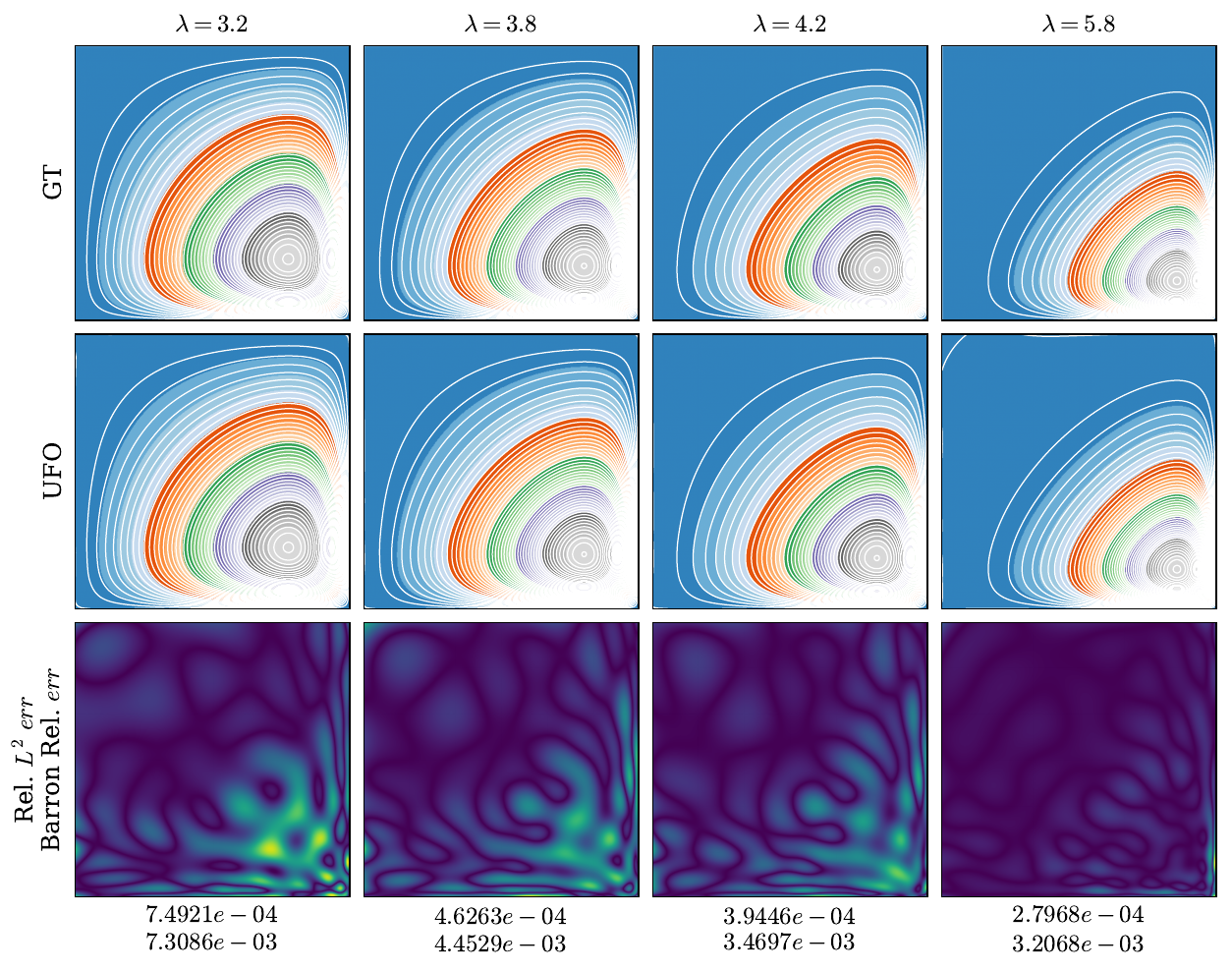}
	\caption{In-distribution predictions on the parametric 2D Burgers equation using UFO.}
	\label{Burgers_UFO}
\end{figure}

\begin{figure}[H]
	\centering
	\includegraphics[width=0.8\textwidth]{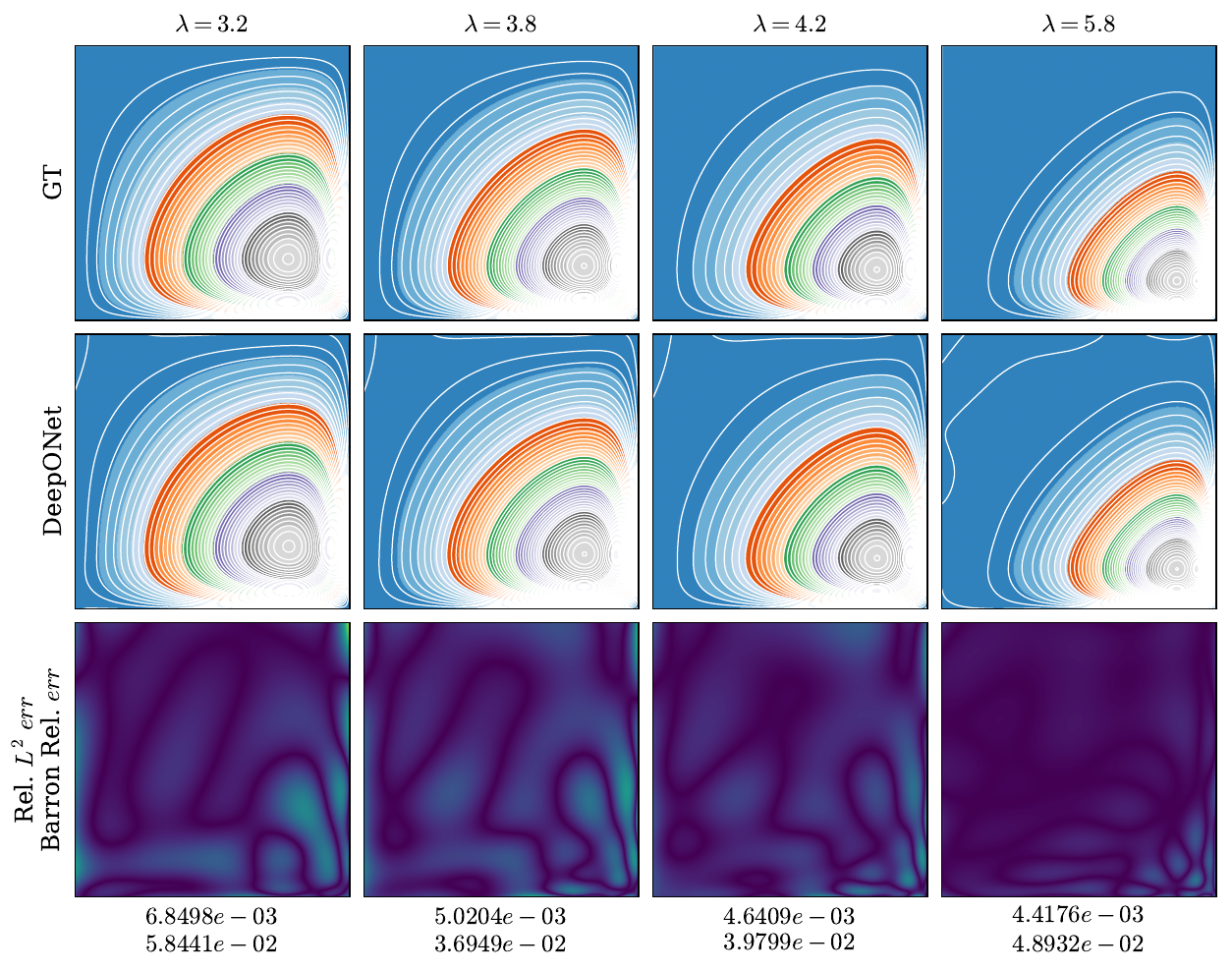}
	\caption{In-distribution predictions on the parametric 2D Burgers equation using DeepONet.}
	\label{Burgers_DeepONet}
\end{figure}

\begin{figure}[H]
	\centering
	\includegraphics[width=0.8\textwidth]{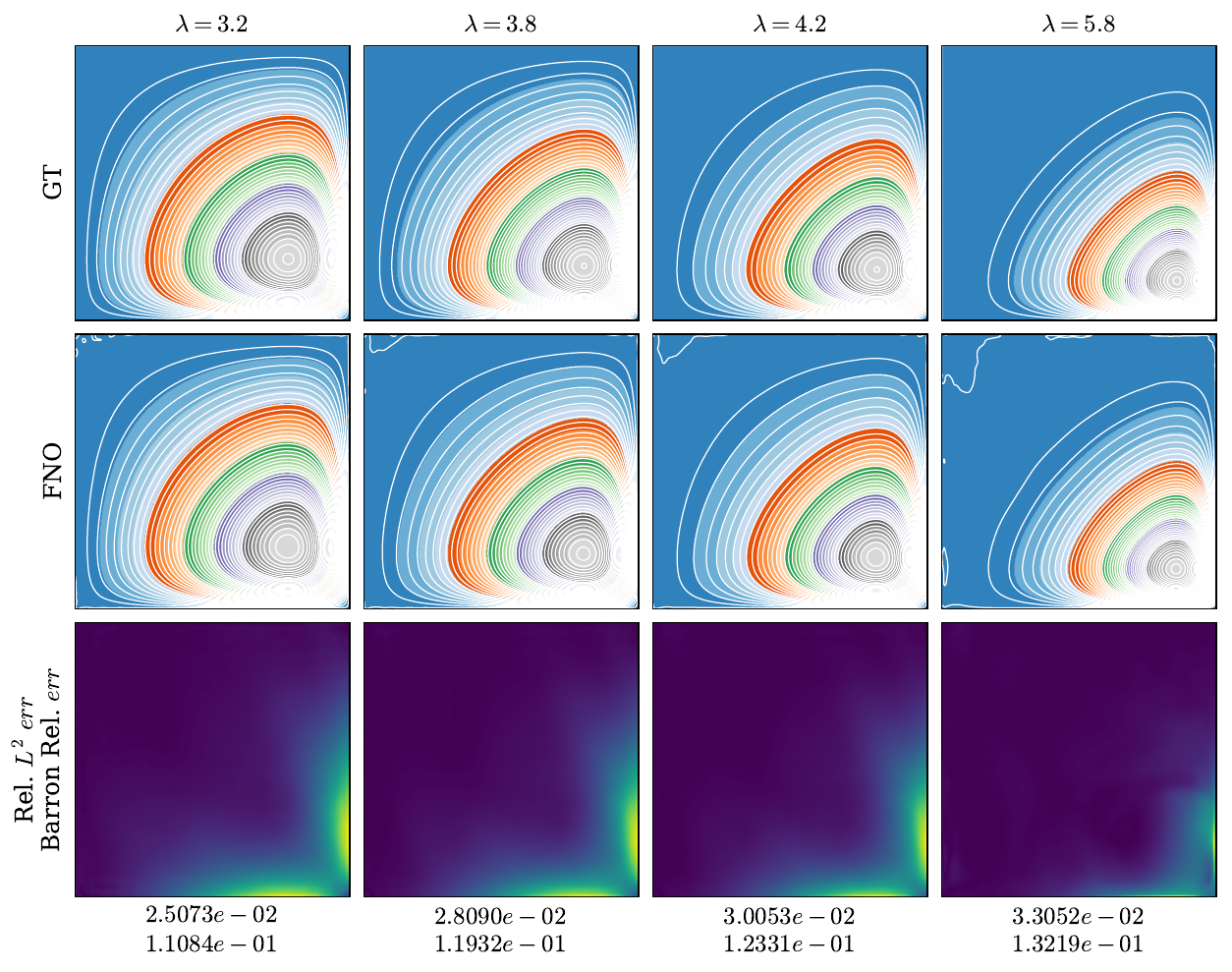}
	\caption{In-distribution predictions on the parametric 2D Burgers equation using FNO.}
	\label{Burgers_FNO}
\end{figure}

\subsection{Additional StepHeat temporal profiles}
Figs. \ref{StepHeat_UFO}-\ref{StepHeat_FNO} provide a profile-level view of the StepHeat predictions corresponding to the quantitative results in Table \ref{tab1}. The ground-truth lines are shown in black. Across different discontinuity locations, all models capture the dominant oscillatory structure and the diffusion-induced amplitude decay over time, confirming that the main temporal dynamics are learned. The remaining discrepancies are concentrated in the high-frequency peaks and troughs, especially at early time snapshots where the discontinuity-induced modes are strongest.

UFO generally tracks the reference profiles closely across the selected \(s\) values, with small deviations in peak amplitude and phase. DeepONet also produces smooth temporal profiles, but shows more noticeable local amplitude mismatch in several cases, consistent with its larger spectral errors in Table \ref{tab1}. FNO captures the periodic structure well in many snapshots, yet exhibits sharper local deviations near high-frequency extrema, reflecting its sensitivity to non-smooth input-induced modes.
\begin{figure}[H]
	\centering
	\includegraphics[width=0.7\textwidth]{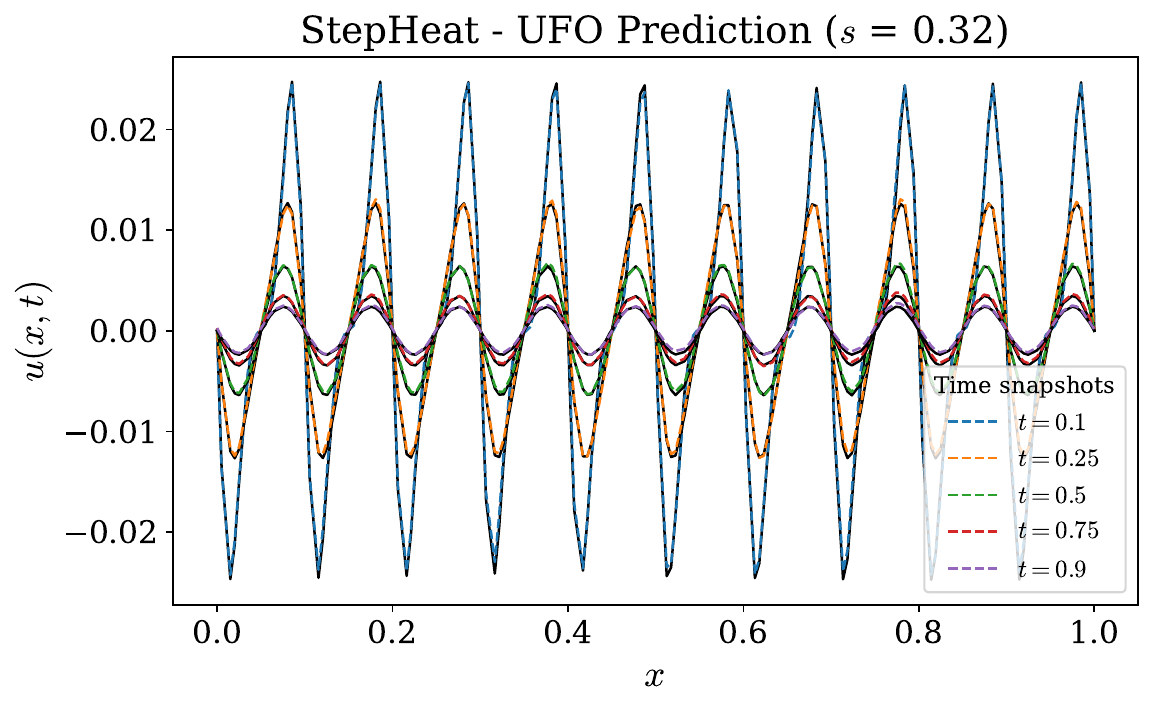}
	\includegraphics[width=0.7\textwidth]{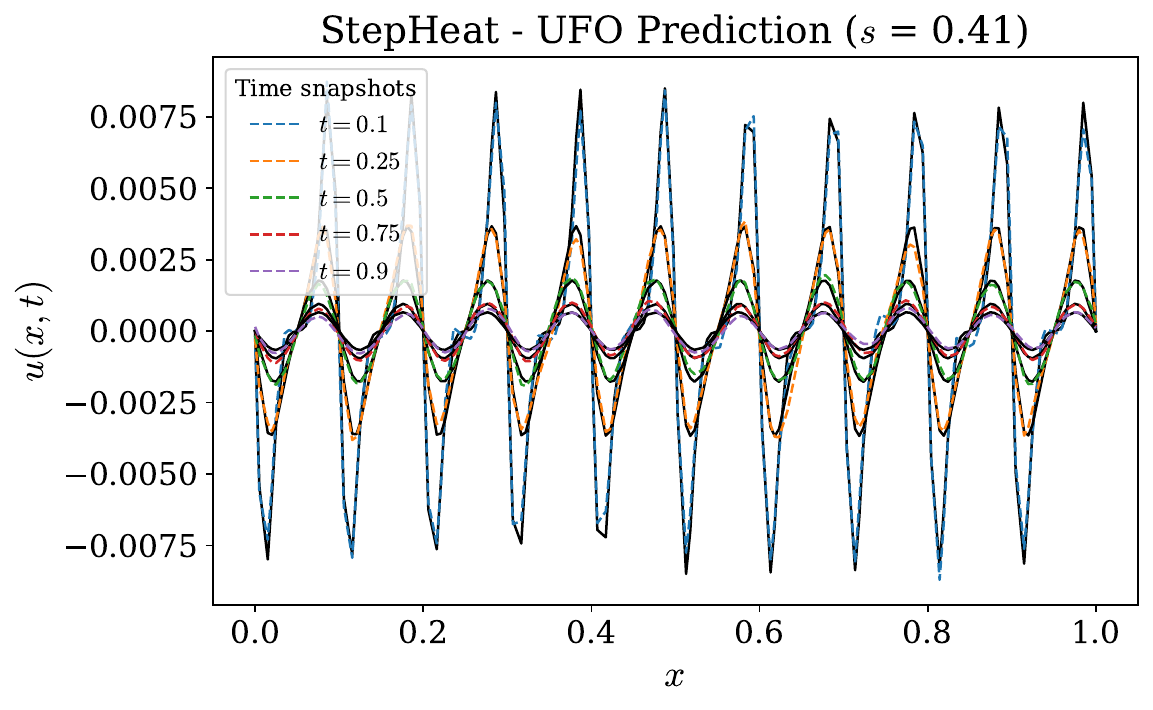}
	\includegraphics[width=0.7\textwidth]{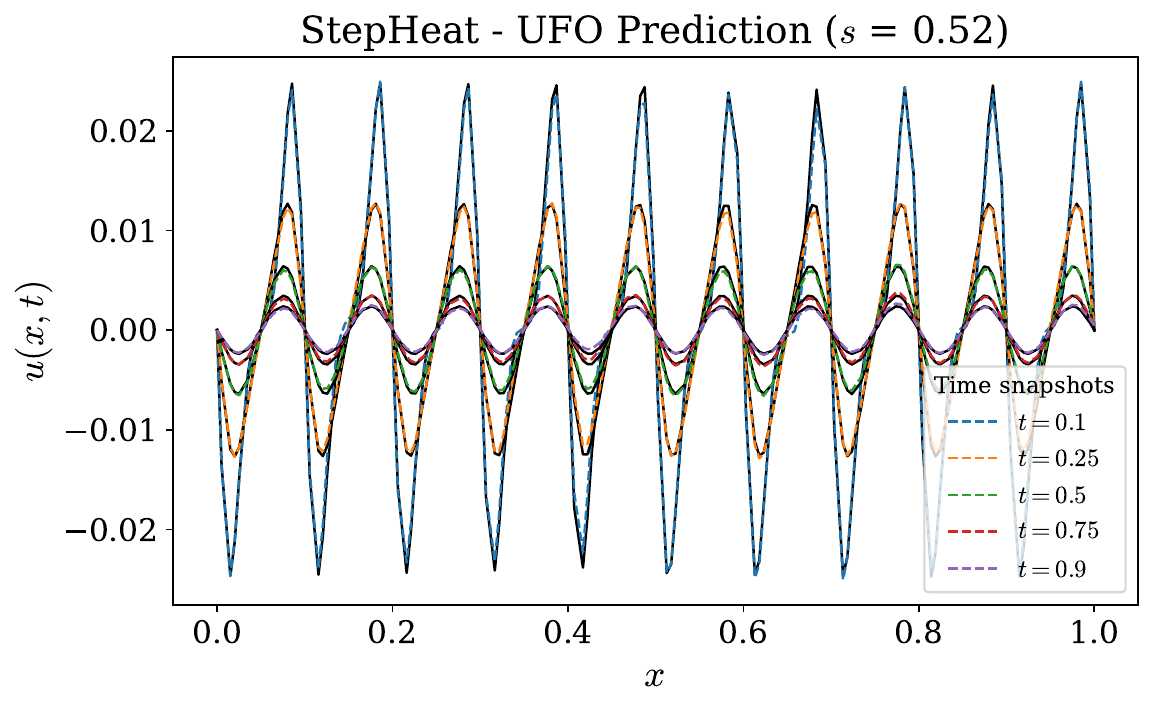}
	\caption{Temporal profile visualization on StepHeat. Predictions are shown for UFO, at \(s=0.32,0.41,0.52\). Each subplot compares the predicted and reference one-dimensional solution profiles at multiple time snapshots \(t\in\{0.1,0.25,0.5,0.75,0.9\}\).}
	\label{StepHeat_UFO}
\end{figure}

\begin{figure}[H]
	\centering
	\includegraphics[width=0.7\textwidth]{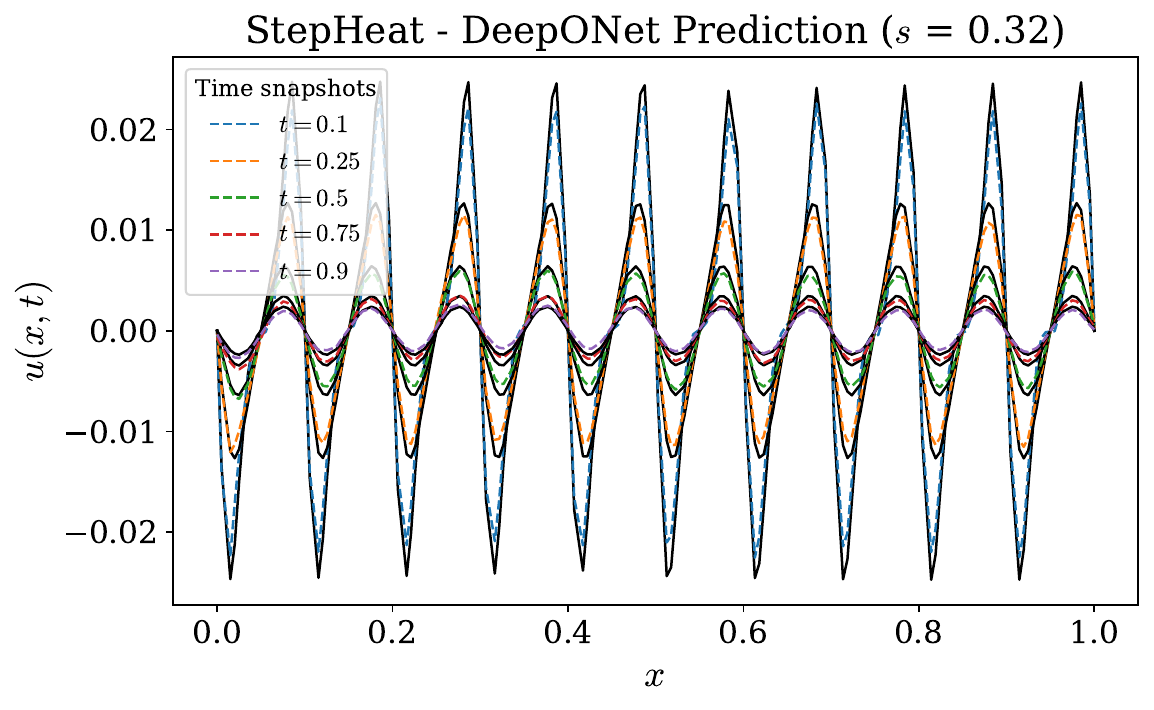}
	\includegraphics[width=0.7\textwidth]{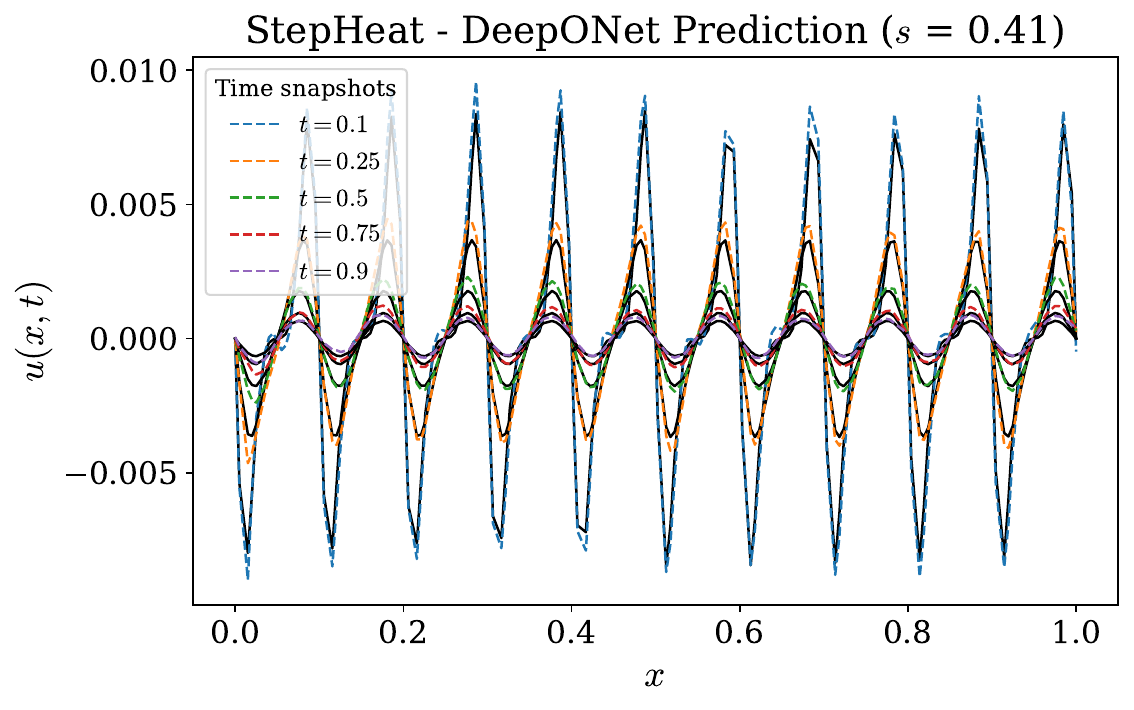}
	\includegraphics[width=0.7\textwidth]{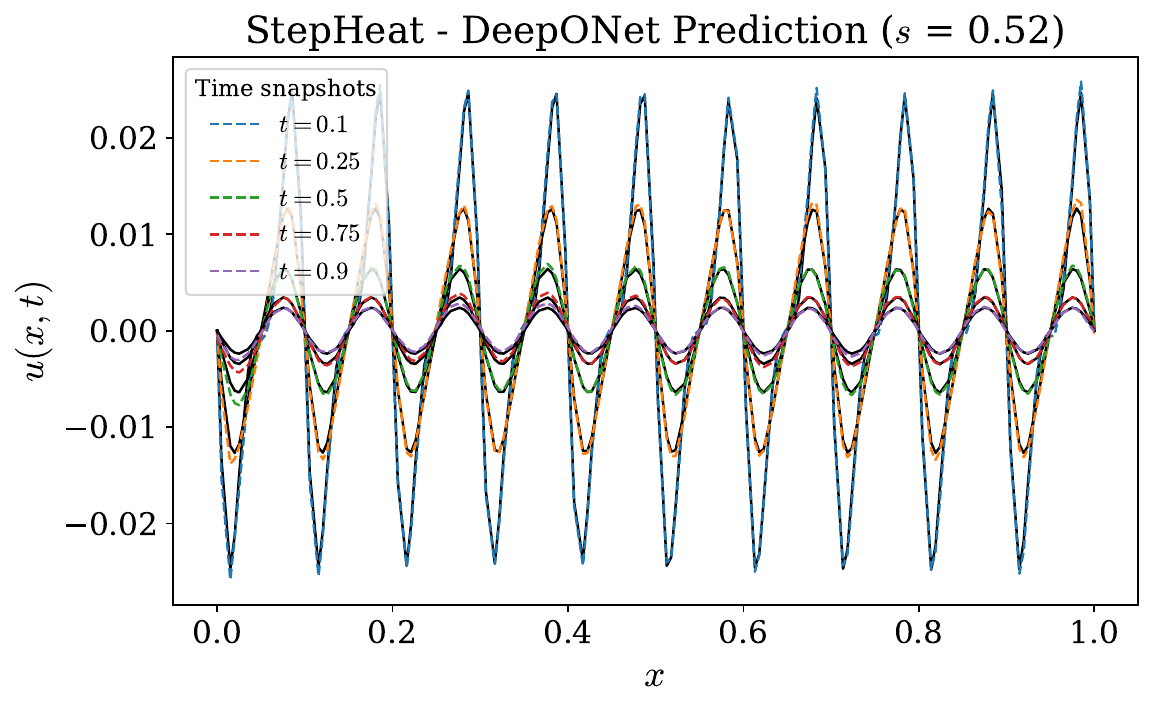}
	\caption{Temporal profile visualization on StepHeat. Predictions are shown for DeepONet, at \(s=0.32,0.41,0.52\). Each subplot compares the predicted and reference one-dimensional solution profiles at multiple time snapshots \(t\in\{0.1,0.25,0.5,0.75,0.9\}\).}
	\label{StepHeat_DeepONet}
\end{figure}

\begin{figure}[H]
	\centering
	\includegraphics[width=0.7\textwidth]{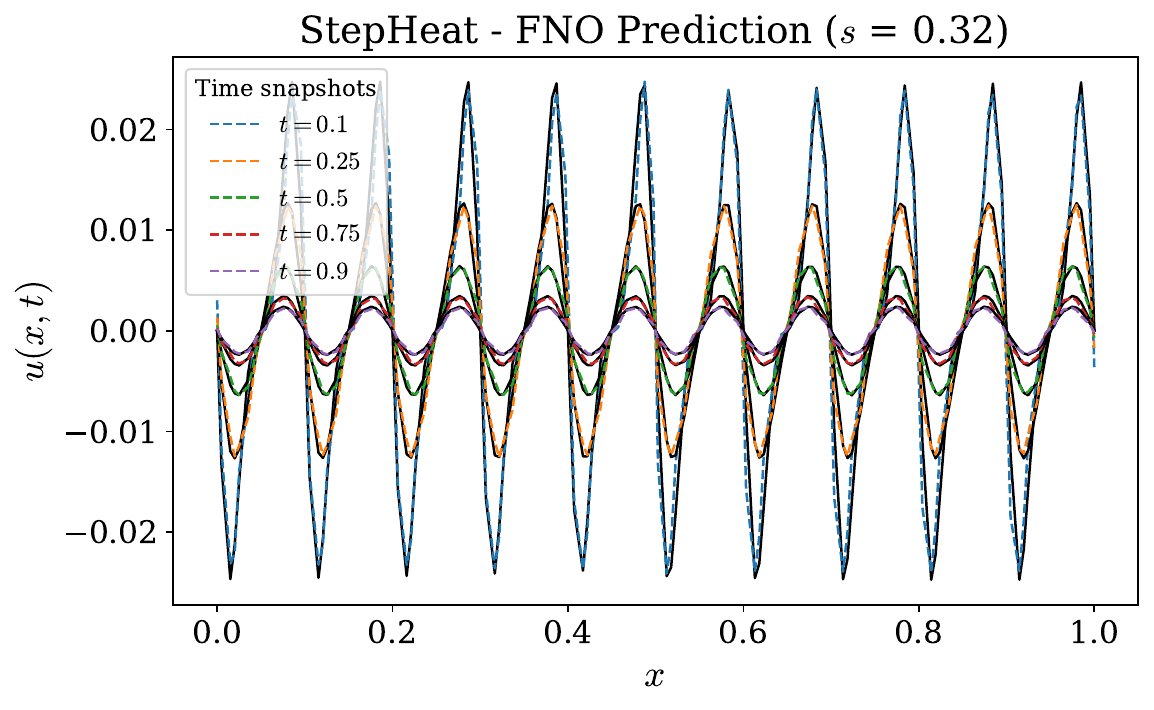}
	\includegraphics[width=0.7\textwidth]{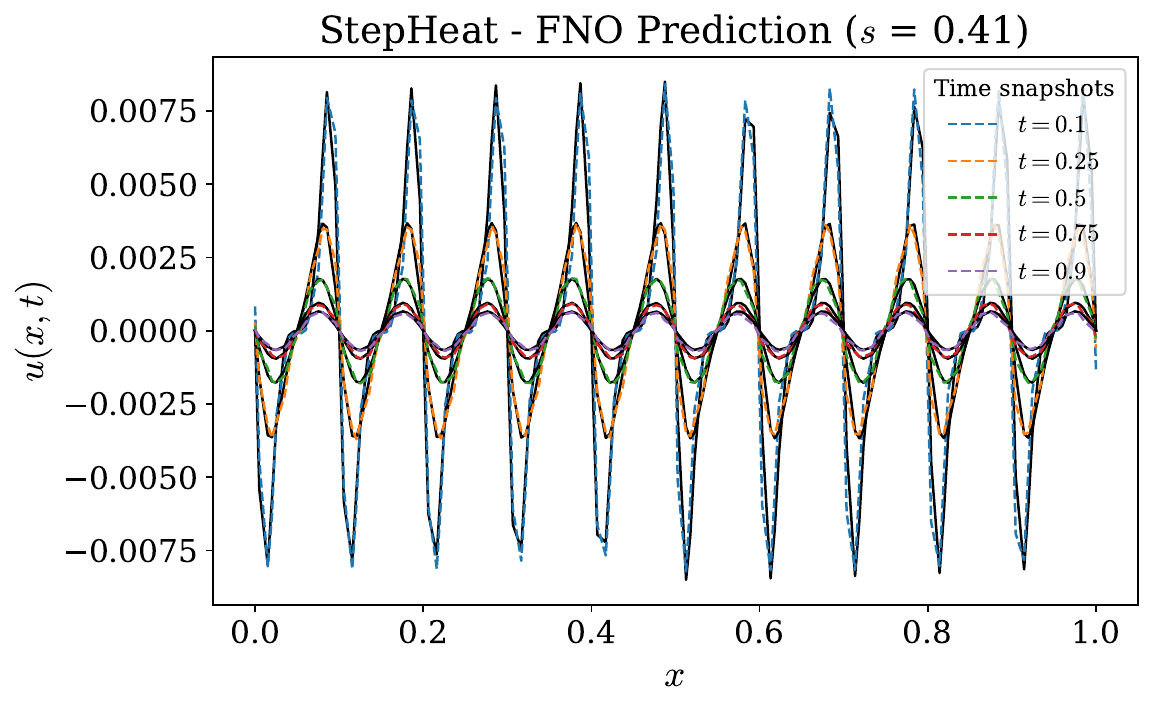}
	\includegraphics[width=0.7\textwidth]{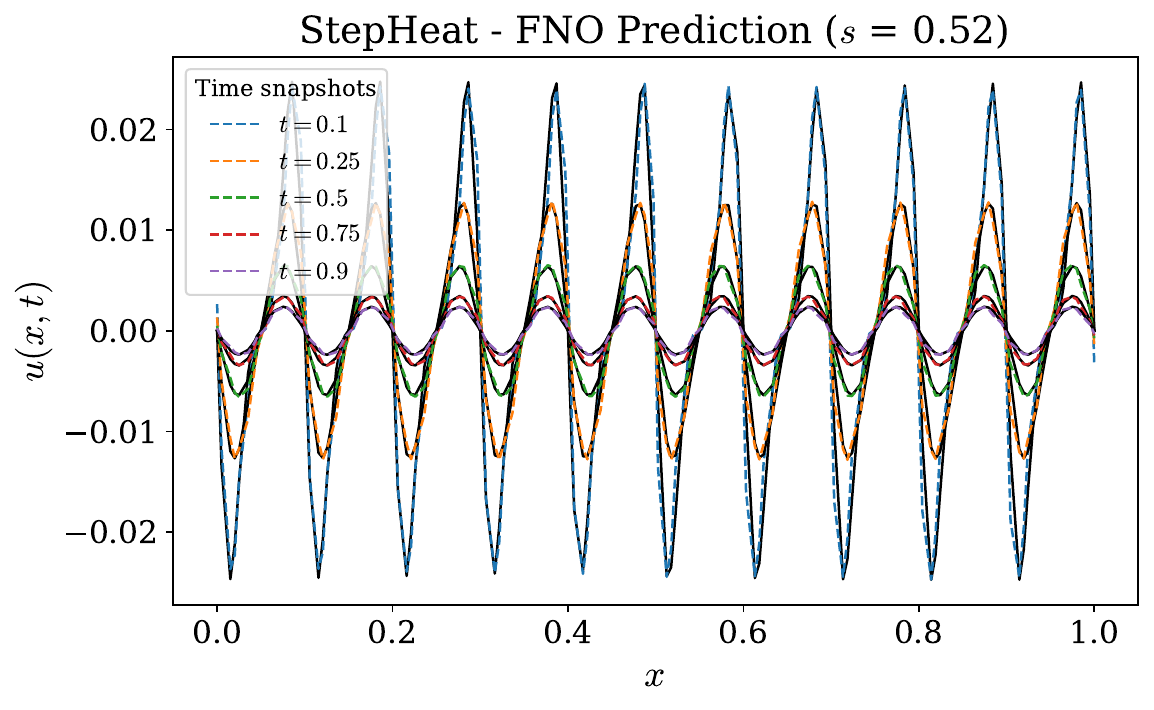}
	\caption{Temporal profile visualization on StepHeat. Predictions are shown for FNO, at \(s=0.32,0.41,0.52\). Each subplot compares the predicted and reference one-dimensional solution profiles at multiple time snapshots \(t\in\{0.1,0.25,0.5,0.75,0.9\}\).}
	\label{StepHeat_FNO}
\end{figure}

\subsection{Additional visualization of \(\delta\)-Helmholtz predictions}
Figs. \ref{delta_UFO}-\ref{delta_FNO} provide additional visualizations with colorbars and absolute error maps. Colorbars are included to expose amplitude shifts and error magnitudes that are not visible from the compact comparison in the main text. The ID case confirms that UFO yields the smallest and most localized residual error, while DeepONet and FNO exhibit more structured error patterns. In the strong OOD cases (\(\delta=\pm30.8\)), the gap becomes more pronounced: UFO preserves the globally shifted oscillatory field with moderate localized errors, whereas DeepONet suffers from structural distortion and FNO often collapses to an incorrect amplitude regime. These results provide a more detailed view of the failure modes behind the compact comparison in the main text. 

\subsection{Additional GRF-Helmholtz samples}
Fig. \ref{GRF_UFO_k60k120} shows additional samples confirm the same trend as discussed in Fig. \ref{GRFk60k120}.
\begin{figure}[H]
	\centering
	\includegraphics[width=\textwidth]{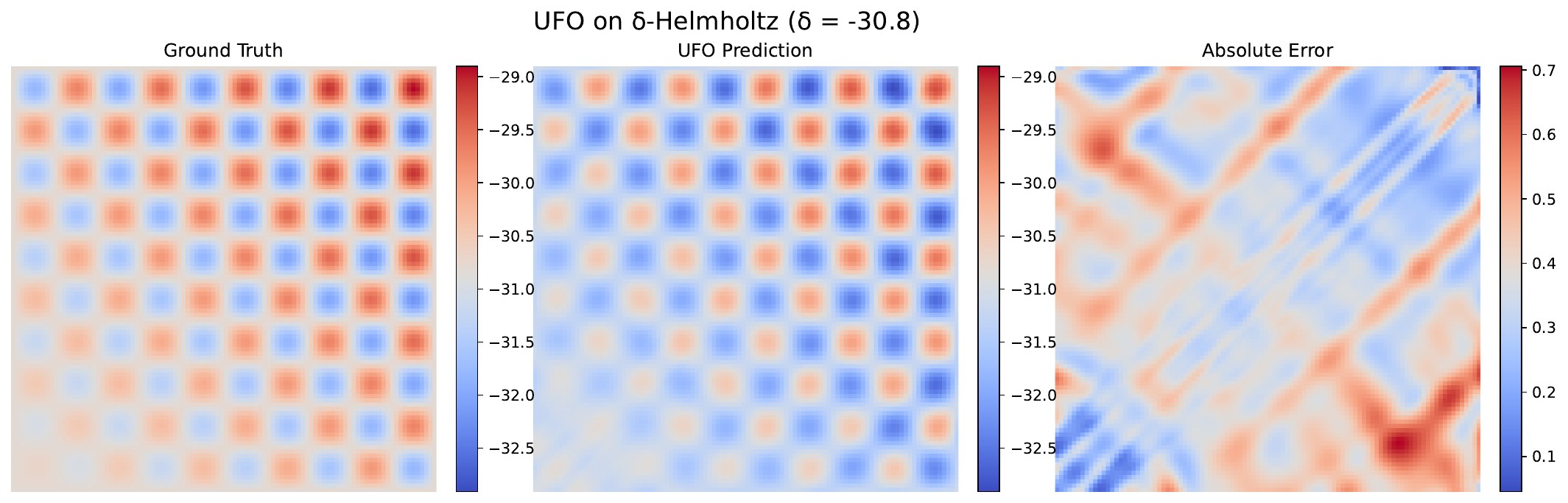}
	\includegraphics[width=\textwidth]{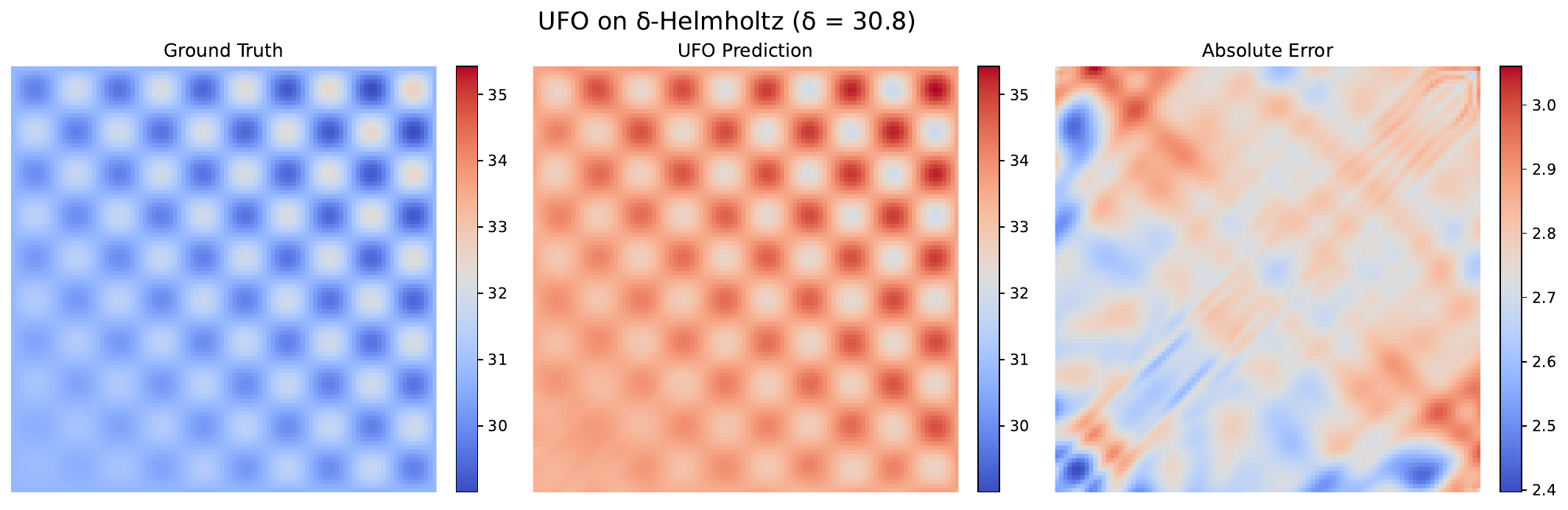}
	\includegraphics[width=\textwidth]{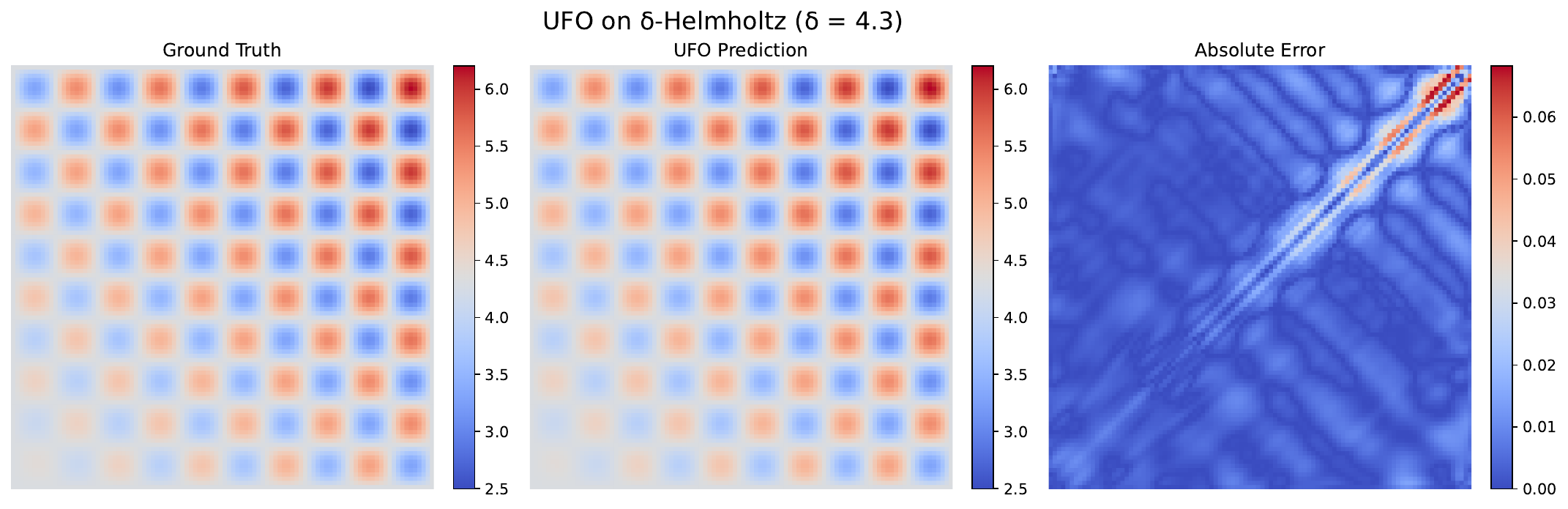}
	\caption{Additional \(\delta\)-Helmholtz visualizations of UFO with colorbars and absolute error maps in the cases corresponding to Fig. \ref{Delta}.}
	\label{delta_UFO}
\end{figure}

\begin{figure}[H]
	\centering
	\includegraphics[width=\textwidth]{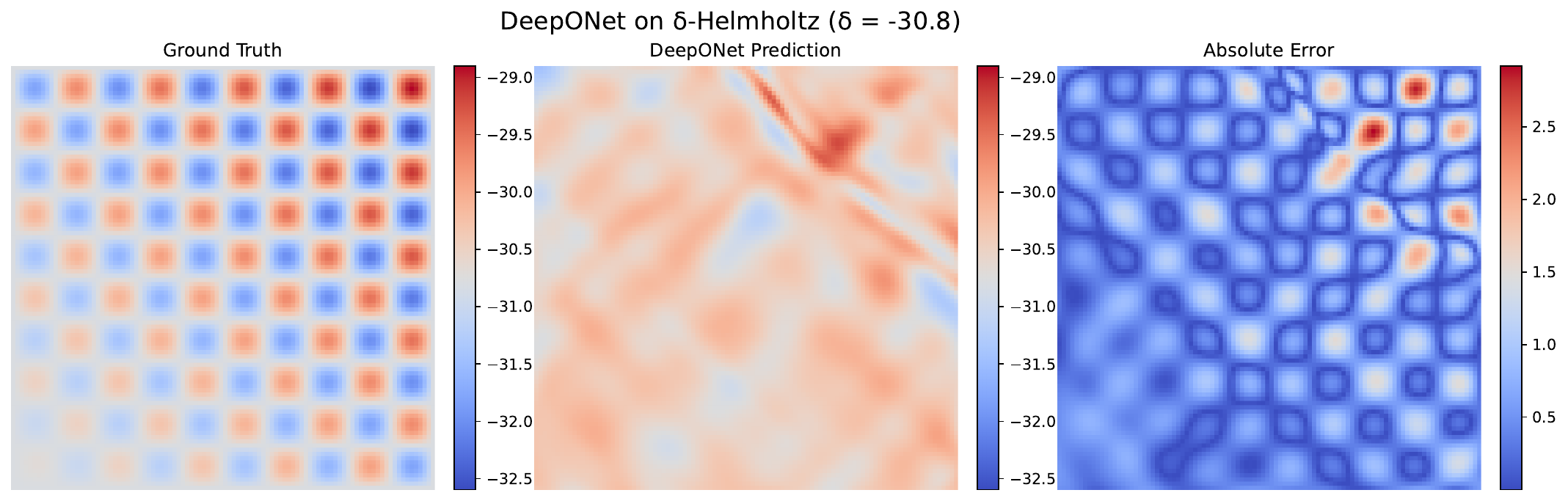}
	\includegraphics[width=\textwidth]{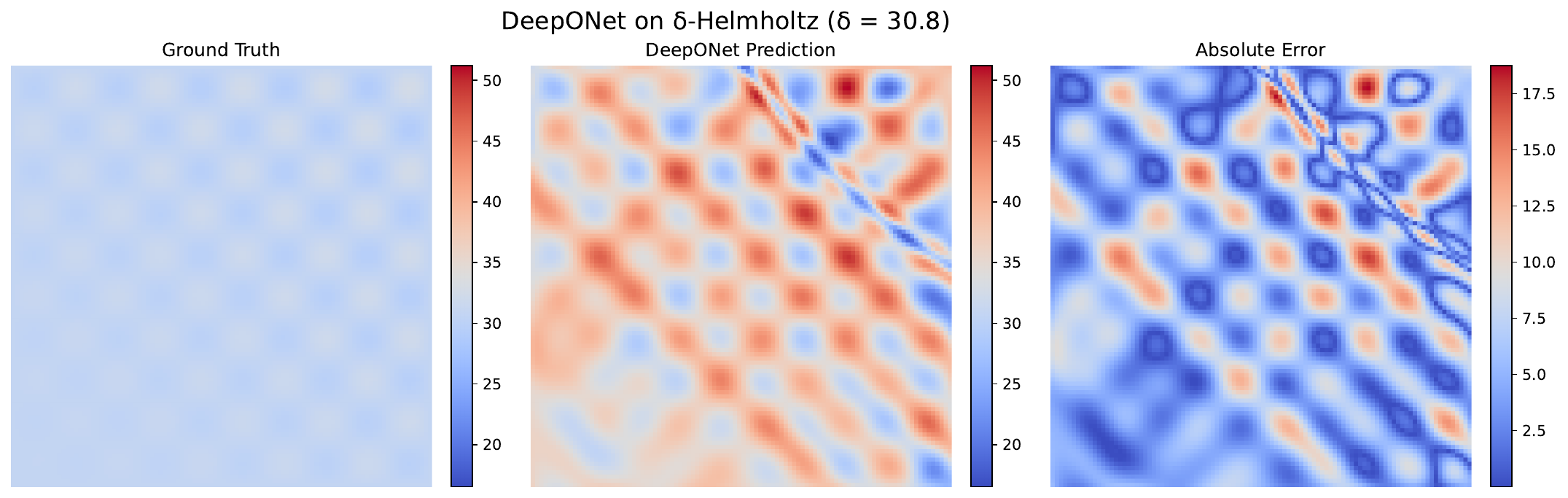}
	\includegraphics[width=\textwidth]{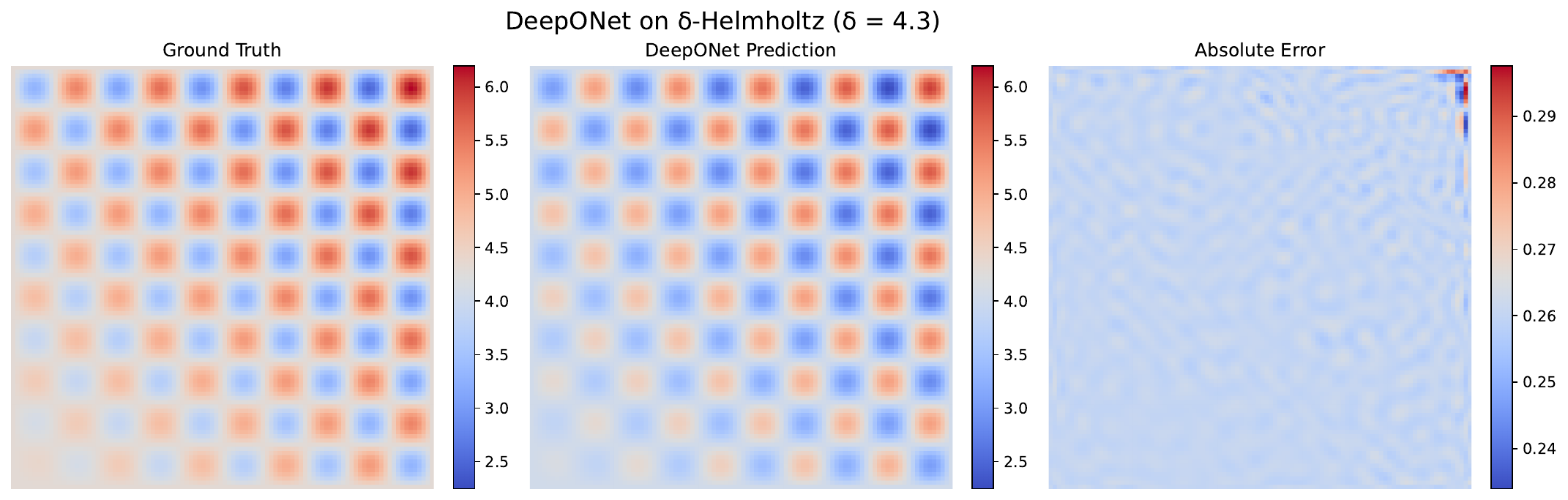}
	\caption{Additional \(\delta\)-Helmholtz visualizations of DeepONet with colorbars and absolute error maps in the cases corresponding to Fig. \ref{Delta}.}
	\label{delta_DeepONet}
\end{figure}

\begin{figure}[H]
	\centering
	\includegraphics[width=\textwidth]{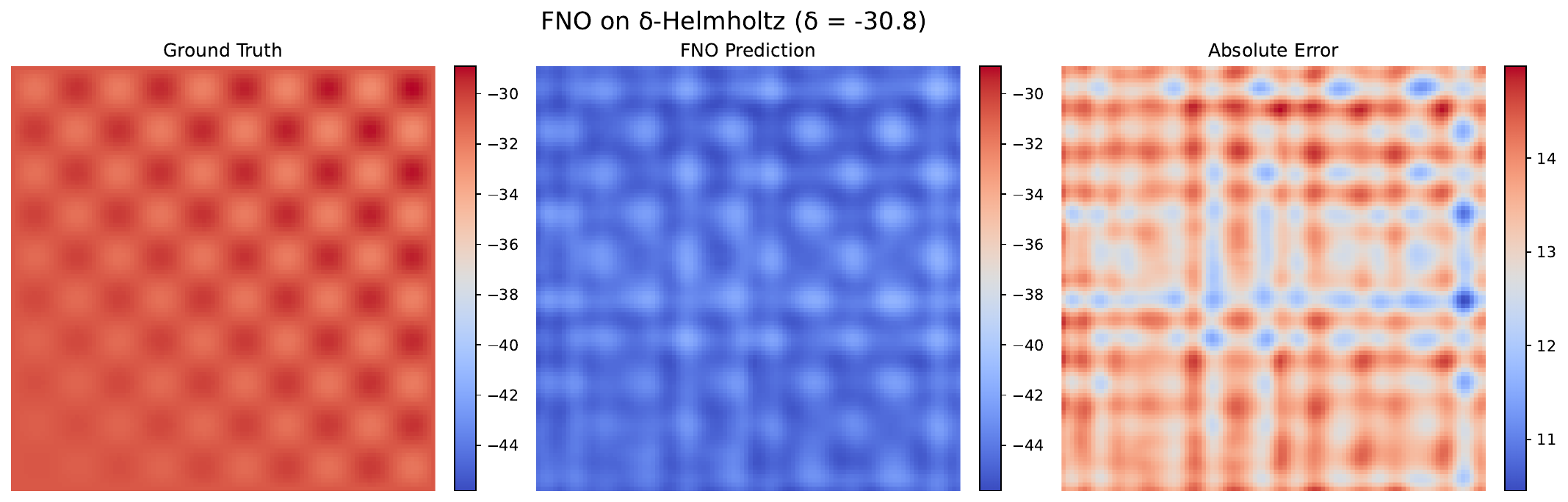}
	\includegraphics[width=\textwidth]{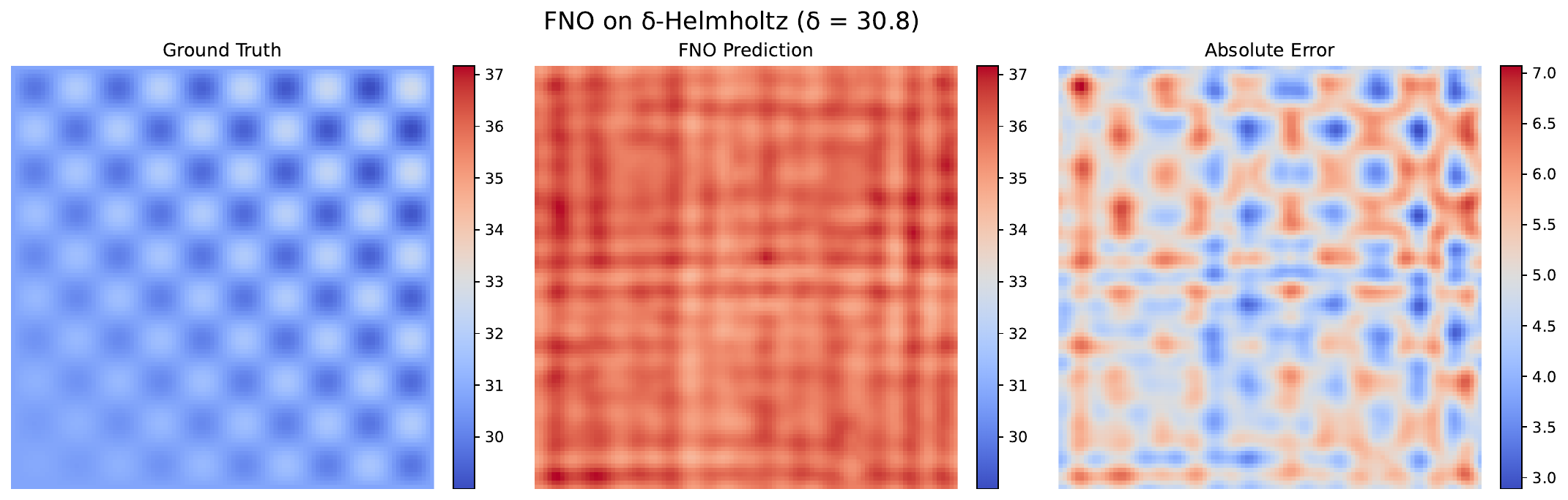}
	\includegraphics[width=\textwidth]{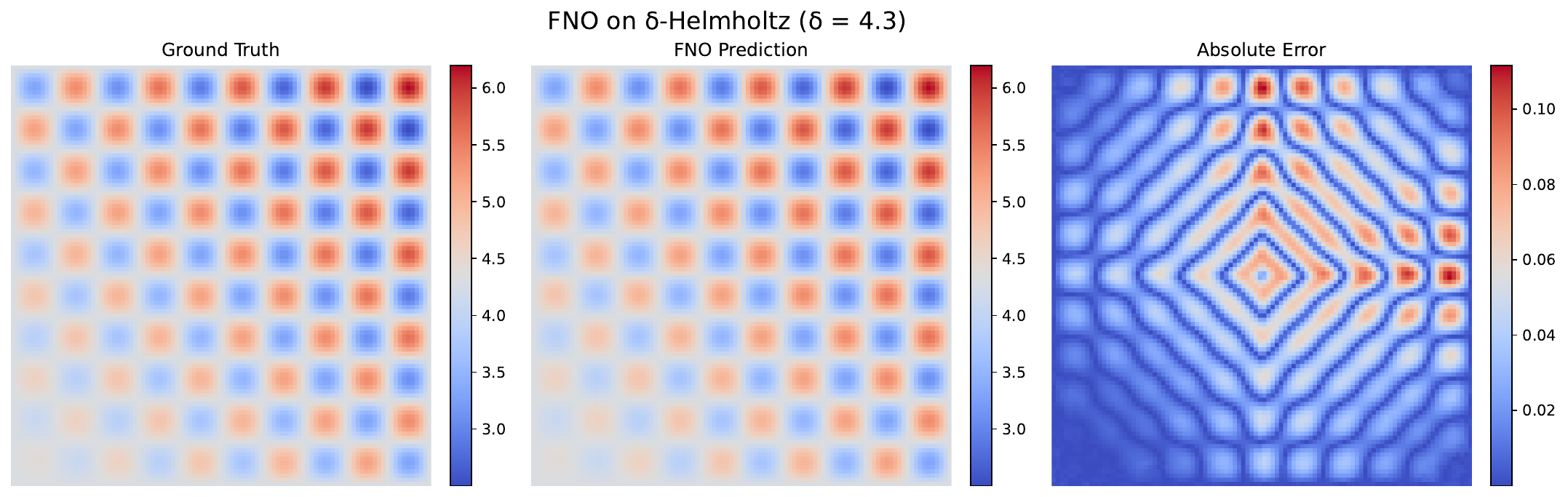}
	\caption{Additional \(\delta\)-Helmholtz visualizations of FNO with colorbars and absolute error maps in the cases corresponding to Fig. \ref{Delta}.}
	\label{delta_FNO}
\end{figure}

\begin{figure}[H]
	\centering
	\includegraphics[width=\textwidth]{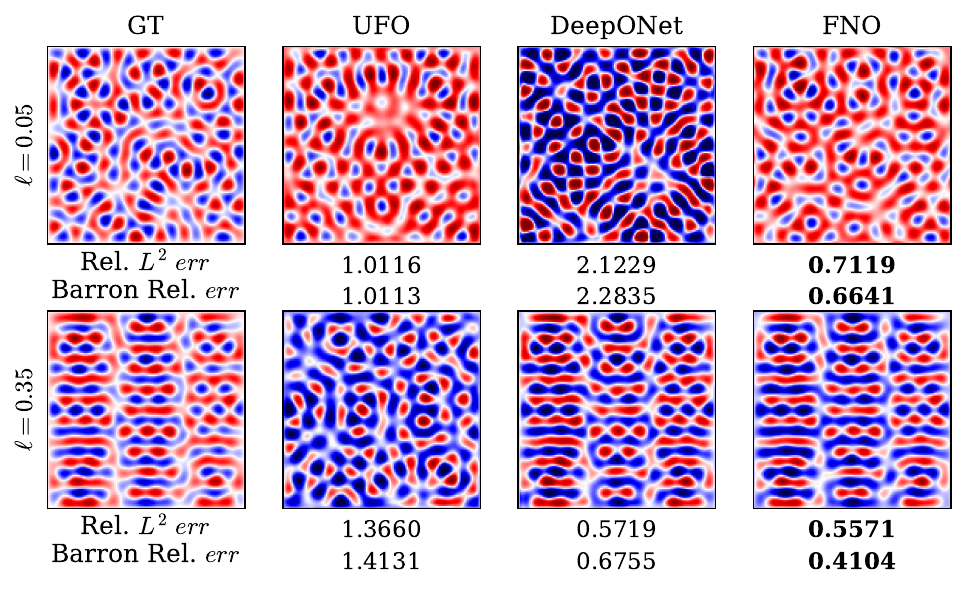}
	\includegraphics[width=\textwidth]{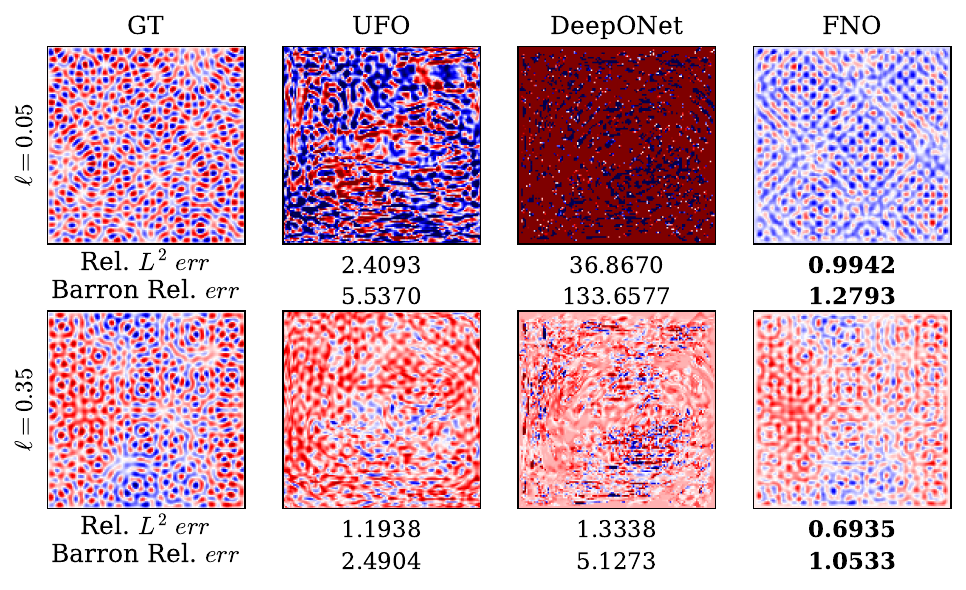}
	\caption{Multiple test samples under OOD correlation lengths \(\ell=0.05\) and \(\ell=0.35\), with relative \(L^2\) and Barron errors reported below each prediction. The top panel presents the predictions at \(k=60\), while the bottom panel shows the results at \(k=120\).}
	\label{GRF_UFO_k60k120}
\end{figure}
\end{bodyspacing}

\end{document}